\pdfoutput=1
\documentclass[11pt]{article}

\usepackage[preprint]{acl}

\usepackage{times}
\usepackage{latexsym}

\usepackage[T1]{fontenc}
\usepackage[utf8]{inputenc}

\usepackage{microtype}

\usepackage{inconsolata}

\usepackage{graphicx}
\usepackage{color}
\usepackage{booktabs}
\usepackage{subcaption}
\usepackage{amsmath}
\usepackage{array}
\usepackage{subcaption}
\usepackage{pdfpages} 

\usepackage{multicol}
\usepackage{makecell}
\usepackage[table]{xcolor}
\usepackage{enumitem}
\usepackage{algorithm}
\usepackage{algpseudocode}
\usepackage{amssymb}
\usepackage{multirow}
\usepackage{soul}
\usepackage{xcolor}

\definecolor{lightblue}{rgb}{0.8,0.9,1} 
\sethlcolor{lightblue}   % set the highlight color

\title{\textsc{Op-Fed}: Opinion, Stance, and Monetary Policy \\ Annotations on FOMC Transcripts Using Active Learning}

\author{Alisa Kanganis\\
  Williams College\\
  \texttt{abk2@williams.edu } \\\And
  Katherine A.~Keith\\
  Williams College\\
  \texttt{kak5@williams.edu} \\}

\usepackage{newunicodechar}
\newunicodechar{↪}{\ensuremath{\hookrightarrow}}

\begin{document}
\maketitle
\begin{abstract}
The U.S.~Federal Open Market Committee (FOMC) regularly discusses and sets monetary policy, affecting the borrowing and spending decisions of millions of people. In this work, we release \textsc{Op-Fed}, a dataset of 1044 human-annotated sentences and their contexts from FOMC transcripts.
We faced two major technical challenges in dataset creation: imbalanced classes---we estimate fewer than 8\% of sentences express a non-neutral stance towards monetary policy---and inter-sentence dependence---65\% of instances require context beyond the sentence-level. 
To address these challenges, we developed a five-stage hierarchical schema to isolate aspects of opinion, monetary policy, and stance towards monetary policy as well as the level of context needed. 
Second, we selected instances to annotate using active learning, roughly doubling the number of positive instances across all schema aspects. Using \textsc{Op-Fed}, we found a top-performing, closed-weight LLM achieves 0.80 zero-shot accuracy in opinion classification but only 0.61 zero-shot accuracy classifying stance towards monetary policy---below our human baseline of 0.89. We expect \textsc{Op-Fed} to be useful for future model training, confidence calibration, and as a seed dataset for future annotation efforts. 
\end{abstract}

\section{Introduction}

The twelve voting members of the United States' Federal Open Market Committee (FOMC) cast votes (approximately) every six weeks to decide the target federal funds rate. These decisions impact hundreds of millions of people---affecting everything from mortgage rates to loan availability---and ultimately influence unemployment and inflation.
Yet, there remain open questions about the internal decision making dynamics of the FOMC, e.g.,~majority dynamics, consensus building, influence of the Chairman, partisan preferences etc. \citep{chappell2004committee}.

Although there is a large body of prior work that analyzes FOMC communications, many empirical studies are limited by their use of dictionary-based or bag-of-words approaches \citep{acosta2015hanging,mazis2017latent,ruman2023comparative,bordo2024rules,aruoba2024long}. Other work is limited by their focus on certain FOMC texts---such as press releases and meeting minutes---which are carefully curated by the FOMC prior to their public release \citep{shah2023trillion}. 

\begin{figure}[t]
\centering
\includegraphics[width=0.98\columnwidth]{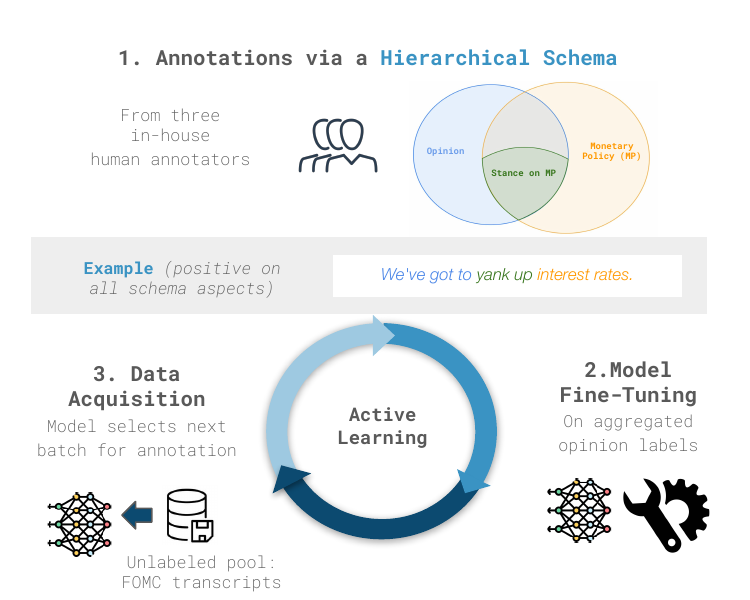}
\caption{\textbf{Overview of \textsc{Op-Fed} dataset creation}.
\textbf{(1)}~In batches, humans annotate sentences from FOMC transcripts using our hierarchical schema. \textbf{(2\&3)}~We fine-tune a pre-trained model on the annotated batch, and acquire a new batch via an AL acquisition function.\label{f:hook}
}
\end{figure}

In contrast, our work focuses on verbatim meeting transcripts---which the FOMC releases after a five-year delay---because studies suggest that the transcripts contain economic signals not present in the FOMC's other released communications.
In particular, \citet{meade2005fomc} find the committee members' formal votes were almost always consensual despite their notable disagreements in the transcripts, and \citet{fischer2023fed} estimated real-time access to FOMC transcripts could substantially increase market forecasting accuracy. 

In an ideal world, economists studying FOMC dynamics would have a (perfectly accurate) classifier which takes as input input an utterance from a FOMC member and outputs the member's stance towards monetary policy. For example, the classifier would output a positive stance towards tightening monetary policy for the utterance \emph{``We've got to yank up interest rates''} and a negative stance for the utterance \emph{``We don't want to keep the funds rate biased upward for too long.''}

Unfortunately, it is currently infeasible to train supervised classifiers or to benchmark the zero-shot classification performance of pre-trained large language models (LLMs) for this task because there does not exist an annotated dataset for individual's stances in FOMC transcripts.\footnote{To the best of our knowledge there are no machine-readable, publicly available datasets for this task.} Thus, the goal of this work is to take the first steps towards operationalizating this downstream task and creating the first dataset for the task. However, in doing so, we faced two major technical challenges.

%Katie: convokit_lookup.py

\textbf{Challenge I: Inter-sentence dependence.} Because the FOMC verbatim transcripts contain rich, extemporaneous discourse, there is often inter-sentence dependence. For example, the sentence  \emph{``Yes, I’m fine with your proposal''}\footnote{Spoken by Thomas~Hoenig, President of Federal Reserve Bank of Kansas, 1991-2011; Transcript ID 20050630 in \citet{chang2020convokit}}  requires the previous 10 utterances and 19 sentences in the conversation to understand \emph{``your proposal''} is co-referent to 
\emph{``the 25 basis point increase in the funds rate''} spoken by another FOMC member. We estimated 65\% of transcript sentences require additional context to classify stances (see \S\ref{subsec:prototype}).
  
\textbf{Challenge II: Imbalanced classes.} In these transcripts, FOMC members spend substantial time discussing topics other than policy actions, e.g., the state of the economy, wording of public communications etc. We estimated fewer than 8\% of all transcript sentences express a non-neutral stance towards monetary policy (see \S\ref{subsec:prototype}). Given a finite annotation budget, uniform random sampling of instances for annotation could yield too few non-neutral instances to benchmark or train a classifier. 

To address these technical challenges, we (1) create a new hierarchical schema to isolate aspects of opinion, monetary policy, and stance towards monetary policy, and (2) deploy \emph{active learning}, a model-and-human-in-the-loop technique that aims to improve on uniform random sampling of unlabeled instances \citep{zhang2022survey}.
Figure~\ref{f:hook} provides an overview of our active learning and annotation pipeline. 
In summary, our primary \textbf{contributions} are:
\begin{itemize}[leftmargin=*,itemsep=1pt, topsep=0pt]
    \item We release \textsc{Op-Fed}\footnote{A word play on ``op-ed'' in newspapers.}, a dataset of 1044 sentences plus their surrounding context that are human-annotated with our hierarchical schema (\S\ref{subsec:schema}) of opinion, monetary policy, and stance towards monetary policy.
    \item We evaluate the zero-shot accuracy of top-performing, closed-weight LLMs (\S\ref{sec:experiments}). The best performing model, \emph{Claude Opus 4.1}, achieves 0.8 accuracy classifying opinion but only 0.61 accuracy for classifying stance towards monetary policy---below our human baseline of 0.89.\footnote{We provide our dataset and code at \url{https://github.com/kakeith/op-fed}.}
\end{itemize}
\noindent
These results suggest \textsc{Op-fed} is a useful dataset for future model training or as a seed dataset for future annotation efforts. 
Next, we describe our data and schema (\S\ref{sec:data-schema}), AL pipeline (\S\ref{sec:al}), and describe how economists might use our dataset in the future (\S\ref{sec:future}). 

\section{Related Work}

\subsection{Analysis of FOMC Communications}

Several studies have examined FOMC text other than transcripts, such as press releases \citep{acosta2015hanging}, FOMC statements \citep{rohlfs2016effects}, speeches \citep{zirn2015lost}, or Federal Reserve staff documents \citep{aruoba2024long}. 
Other economists have attempted to analyze transcripts manually---e.g., \citet{chappell2004committee} inferred individual members' policy reaction functions based on their 
stated preferences for the federal funds rate in the transcripts. However, these manual analyses are not in a machine-readable format. 
Other studies have applied NLP to FOMC transcripts using simple dictionary-based or bag-of-word approaches \citep{mazis2017latent,ruman2023comparative,bordo2024rules}. 

The two most closely related works to ours are \citet{shah2023trillion} and \citet{peskoff2023gpt}.
\citeauthor{shah2023trillion}~created an annotated dataset of FOMC speeches, minutes, and press conference transcripts for hawkish-dovish classification. However, they use keywords (e.g., ``inflation'' and ``fund rate'') to filter the initial dataset, resulting in keyword bias. Additionally, they operationalize their \emph{dovish} label as ``any sentence that indicates future monetary policy easing'' which leaves large implicature gaps, e.g., \emph{``In addition, U.S. inflation remains muted''} is classified as dovish. 
Using a small manually-annotated sample, \citet{peskoff2023gpt} estimate an F1 score of 0.57 for zero-shot predictions from LLMs (GPT-3 and GPT-4) on the same hawkish-dovish classification task. 
\textsc{Op-Fed} differs from these existing annotated FOMC transcript datasets in several key ways: (1) we incorporate context beyond the sentence-level, (2) we address class imbalance by deploying active learning, and (3) we annotate the personal stances of individual members to enable more fine-grained downstream inference. 

\subsection{Opinion Mining \& Stance Detection}\label{sec:related-stance}

Our schema draws inspiration from the tasks of opinion mining \citep{pang2008opinion,stoyanov2005multi,wilson2017mpqa} and stance detection.
Stance detection aims to identify speakers' positions toward a specific target, claim, or proposition, and stance datasets have been released for a variety of domains including U.S.~politics \citep{mohammad-etal-2016-semeval}, companies' mergers and acquisitions \citep{conforti-etal-2020-will}, and judicial opinions \citep{gupta2025stance}. 
As \citet{allaway2020zero} discuss, there are two different operationalizations of stance detection.
In the most common operationalization---\emph{topic-phrase stance}---a text is classified with stance labels \{\emph{pro, con, neutral}\} towards a topic which is typically a noun-phrase, e.g., ``gun control'' \citep{kuccuk2020stance}. In the second
operationalization---\emph{topic-position stance}---stance with labels  \{\emph{agree, disagree, discuss, unrelated}\} is detected between
a text and a topic that is an entire position statement, e.g., \emph{``We should disband NATO''}. Our hierarchical schema builds upon this later topic-position stance operationalization (\S\ref{subsec:schema}). 

\subsection{Active Learning for NLP}

There is a large literature that combines active learning (AL) with NLP; see \citet{settles2009active,zhang2022survey} for surveys. Previous work has proposed NLP-specific AL strategies such as using the pre-training loss as a proxy for uncertainty \citep{yuan2020cold}, using AL after domain-adaptive pre-training \citep{margatina2022importance}, or using parameter efficient fine-tuning (PEFT) within the AL loop \citep{jukic2023parameter}. Most closely related to our AL setting, \citet{dor2020active} found active learning with BERT models outperformed randomly sampling by the largest margin for datasets with imbalanced class distributions.
One possible concern of using AL is that ``training a successor model with an actively-acquired dataset does not consistently outperform training on i.i.d.~sampled data'' \citep{lowell2019practical}. However, \citet{margatina2023limitations} argue that AL evaluations with simulated data from clean, curated datasets---which typically have balanced class proportions---may only be a \emph{lower} bound to their effectiveness on real (noisy, class-imbalanced) data.
\section{Data \& Annotation Schema}\label{sec:data-schema}

\subsection{Data \& Pre-processing}

We use the \emph{FOMC Corpus} provided by the Cornell Conversational Analysis Toolkit \citep{zhang2018convokit}. This is comprised of meetings held between January 1977 and December 2008 and totals 286 transcripts and 364 distinct speakers. We used spaCy for sentence segmentation \citep{honnibal-johnson-2017-spacy}. We excluded sentences from utterances with fewer than four white-space tokens (19,221 total) as these were often repetitive, discourse cues such as \emph{``thank you''}. We also removed utterances---a single turn of a speaker in a conversation---exceeding 500 white-space tokens in length (3,955 total) as these were  almost always factual briefings about economic conditions rather than substantive policy discussions. After these filters, we retained 85,328 utterances (280,975 sentences). 

\begin{table*}[t]
\renewcommand{\arraystretch}{1.3}
\centering
\resizebox{0.98\linewidth}{!}{
\begin{tabular}{>{\raggedright\arraybackslash}m{0.70\textwidth}
c c c c c 
}
  \toprule
  \toprule
  & \multicolumn{5}{c}{\textbf{Schema Aspect}} \\
  \cline{2-6}
      \textbf{Sentence} & Opinion & MP & MP Ctxt. & StanceNLI & StanceNLI Ctxt. \\
     \toprule
     \toprule
     %19811222_230_4
     \textit{``But the discrepancies in the numbers were even larger than we had.''} & no & * & * & * & *
     \\
     \hline 
     %9910206_54_6
     \textit{``If the Congress feels comfortable with \hl{the arrangement}, I'm certainly on board in this context.''} & yes & no  & * & * & * \\
     \hline
     %19810203_410_1
     \textit{``I don't know how important \hl{all of this} is.''} & 
     yes & yes & -5 sentences & neutral & sentence
     \\
     \hline
     %19791006_304_14
     \textit{``\hl{That} sounds like a good idea.''}
     & yes & yes & utterance & contradiction & utterance 
     \\
    \hline 
     \textit{``I would go with \hl{"B"} also.''} &
     yes & yes & -5 sentences & entailment & -5 sentences
     \\
     \hline
     \textit{``Well, I'm clearly not in tune with the other members of the FOMC.''} & yes & yes & utterance & entailment & utterance \\
      \bottomrule
      \bottomrule
    \end{tabular}
  }
  \caption{
  \textbf{Challenging examples from \textsc{Op-Fed} and their gold-standard labels.} 
  We hand-selected shorter sentences (for display reasons) and \hl{highlighted phrases} that require context beyond the sentence-level to be resolved; their full context is in Table~\ref{tab:example_annotations-context}.
  Here, $*$ indicates the label is not applicable due to the annotation hierarchy. 
  }
  \label{tab:example_annotations}
\end{table*}

\subsection{Five-Stage Hierarchical Schema}\label{subsec:schema}

A motivating downstream application for our work is an economist who aims to measure when FOMC members express a directional stance towards monetary policy---specifically, whether a speaker supports contractionary actions (e.g., raising rates) or expansionary actions (e.g., lowering rates) by the Federal Reserve. 

To develop an annotation schema, we (the authors) first conducted several rounds of qualitative exploration of the data via uniform random sampling of sentences. 
However, we found it difficult to simultaneously assess all aspects needed to classify a speaker's stance. We decided a 
\textbf{hierarchical annotation schema} could potentially reduce annotator burden and isolate individual aspects for a model in our AL loop. Hierarchical annotation schemas have shown to be useful in other NLP tasks, e.g., annotating clinical notes \citep{gao2022hierarchical}, argumentative discourse structures \citep{stab2014identifying}, and ideology detection \citep{liu2024encoding}.
Second, we found many sentences required inter-sentence coreference resolution (\textbf{Challenge I}); 
% such as ``Mr. Chairman, I support \emph{your recommendation}'' and ``Yes, I’m fine with \emph{your proposal}''; 
see Table~\ref{tab:example_annotations} for examples. 
Short responses such as ``\emph{Me too}'' also lacked sufficient context for labeling. Thus, we included in our schema indications of the amount of context required for labels.

To operationalize stance, we use natural language inference (NLI), the task of determining the logical relationship between a premise sentence and hypothesis sentence \citep{bowman2015large,williams2017broad}. We use the hypothesis---equivalent to the the ``stance position statement'' (\S\ref{sec:related-stance})---``\emph{We should tighten monetary
policy}'' because an \emph{entailment} label would capture all three aspects of interest: stance, monetary policy and tightening direction (e.g., raising target rates). A \emph{contradiction} label would instead capture a stance direction towards loosening monetary policy (e.g., lowering target rates).
 NLI has been used by other computational social science work to classify events in news articles \citep{halterman2021corpus} and political texts \citep{Burnham2025}.

In summary, our schema has annotators proceed hierarchically through the following five stages: 

\begin{enumerate}[leftmargin=*,itemsep=1pt, topsep=0pt]
    \item \textbf{Opinion:} Annotators first determine whether the speaker of the sentence expresses an \emph{opinion} (towards any subject), which we define as \emph{the expression of a subjective perspective, recommendation, or position.} Possible labels are \textit{yes}, \textit{no}, and \textit{ambiguous}.
    
    \item \textbf{Monetary Policy (MP):} If an opinion is present, the annotators then assess whether the instance referenced monetary policy tools or actions (e.g., the federal funds rate or the money supply). Possible labels are \textit{yes}, \textit{no}, and \textit{ambiguous}.
    
    \item \textbf{MP Context:} If the MP label was positive, annotators record how much contextual information is required to determine the label. Possible labels are \textit{sentence}, \textit{utterance}, \textit{-5 sentences} (previous five sentences), and \textit{-200 tokens} (previous two hundred tokens, rounding up to the nearest full sentence).
    
    \item \textbf{StanceNLI:}
    Given \emph{yes} labels for Stages 1 and 2, annotators then choose whether the target sentence \emph{entails}, \emph{contradicts}, is \emph{neutral}, or is \emph{ambiguous} in reference to the hypothesis statement \textit{``We should tighten monetary policy.''}

    \item \textbf{StanceNLI Context:} Annotators then indicate the amount of context required for the NLI label. Possible labels are the same as Stage 3.
\end{enumerate}

\subsection{Prototype Annotation Round}\label{subsec:prototype}
To estimate class proportions, 
we sampled 200 sentences uniformly at random from our pool of transcript sentences, and one of the authors annotated the sentences given our five-stage schema.
We found for Stage 3, 54\% of instances required additional context (beyond the sentence-level) to determine MP, and for Stage 5, 65\% required additional context for the NLI label, confirming \textbf{Challenge I}. 
We also observed the following (second column; Table~\ref{tab:descriptive}): for Stage 1, 24.5\% of the sentences exhibited an opinion; for Stage 2, 13.5\% contained both an opinion and a reference to MP; and for Stage 4, only 7.5\% expressed a non-neutral stance on MP.

\section{Active Learning Pipeline}\label{sec:al}

\subsection{AL for \textsc{Op-Fed} Set-Up}\label{subsec:al-decisions}
Given the extreme class imbalance observed in the prototype annotation round (\textbf{Challenge II}) and our limited annotation budget, we decided to deploy active learning (AL) with the goal of increasing our sampling of positive instances for annotation. 
We decided the classifier within our AL loop would only model Stage 1 (opinions) for several reasons. First, we observed in our prototype annotation round that all opinion labels could be inferred using sentences alone (avoiding \textbf{Challenge I}), which increases the chance of AL being successful since sentence-level models typically perform better than longer context models \citep{mishra2021looking,zhang2024found}. Second, we hypothesized increasing the sampling of positive instances in the first stage in our hierarchical schema could increase positives in all subsequent aspects (a hypothesis we empirically confirmed to be correct; see Table~\ref{tab:descriptive}). We leave to future work using multi-task learning of all stages within the AL loop; see~\citet{rotman2022multi}. 

\subsection{AL Overview}

We implement pool-based, cold-start active learning. Following the definition of active learning provided by \citet{margatina2023limitations}, 
let $C$ be a corpus of text. Let ${A} \subset {C}$ and ${B} \subset {C}$ both be labeled datasets and $|{A}| = |{B}|= \mathcal{D}$ where $\mathcal{D}$ is the annotation budget. We denote a model class $M$ and the data on which it has been trained with subscripts $M_A$ and $M_B$. 
Let ${T} \subset {C}$ be a labeled test set. 
We say $A$ is a ``better'' dataset if $\delta(M_A (T)) > \delta(M_B(T))$ for the chosen performance metric $\delta$. Otherwise, $B$ is better. 
In pool-based active learning, the model $\mathcal{M}$ is first trained on the initial subset of labeled data $\mathcal{D}_{\text{lab}}$. Following prior work, our annotators label a uniform random sample for the initial labeled set. Then, using the model in-the-loop, an acquisition function $A$ samples a batch $\mathcal{Q} \subset \mathcal{D}_{\text{pool}}$ of the ``best'' sentences in $\mathcal{D}_{\text{pool}}$. Here, ``best'' is typically a trade-off between informativeness (e.g., selecting instances with the highest uncertainty or disagreement) and representativeness (e.g., selecting instances that are different from already labeled instances). $\mathcal{Q}$ is subsequently labeled by human annotators, appended to $\mathcal{D}_{\text{lab}}$, and removed from $\mathcal{D}_{\text{pool}}$. We then re-train the model $\mathcal{M}$ on the updated $\mathcal{D}_{\text{lab}}$ and we record its performance on a held-out test set $\mathcal{D}_{\text{test}}$.

\textbf{AL Acquisition Functions.}
We implement and experiment with several acquisition functions that trade-off informativeness and representativeness. 
As our baseline for representativeness, we use \textbf{n-gram density} which calculates the number of unseen n-grams, normalized by the sentence length \citep{eck2005low}. For informativeness, we implement \textbf{maximum entropy} \citep{schroder2021revisiting} which selects ``uncertain'' instances, those for which the model's predictive distribution is closer to uniform. Additionally, we implement \textbf{contrastive active learning (CAL)} \citep{margatina2021active}---a hybrid approach that targets ``contrastive samples'', i.e.~instances that are similar in the model feature space and yet close to the model decision boundary. See \S\ref{asec:aquisition} for more details. 

\subsection{Models}
Within the AL loop, we fine-tune ModernBERT with our opinion (Stage 1) annotations. 
ModernBERT has outperformed other competitive encoder-only models on benchmarks like GLUE while being faster and more memory efficient at inference time \citep{warner2024modernbert}.\footnote{In our pipeline, we use open-weight, encoder-only pre-trained models. Open-weight models are necessary because we need to update the model weights within the AL loop. We use encoder-only models because many of the AL acquisition functions require predicted probabilities and extracting probabilities from generative, decoder-only models requires extensive prompt engineering; see \citet{tian2023just}}
In each AL iteration, we fine-tune the model's weights from its pretraining checkpoint. 
We fix all hyperparameters because at true inference time---when we deploy the AL loop for annotators---we do not have a separate validation set so searching over hyperparameters is infeasible. Other simulation AL work found ``ignoring the dev data and setting a constant number of epochs yielded qualitatively similar, albeit noisier, results'' \citep{dor2020active}.
See Table~\ref{tab:hyperparams} for the full list of hyperparameters.
In our simulations, we also compare ModernBERT to RoBERTa \citep{liu2019roberta} and a baseline of bag-of-words logistic regression (see \S\ref{asec:results}).

\begin{figure*}[t!]
  \centering
  \begin{subfigure}[t]{0.5\textwidth}
    \includegraphics[width=\textwidth]{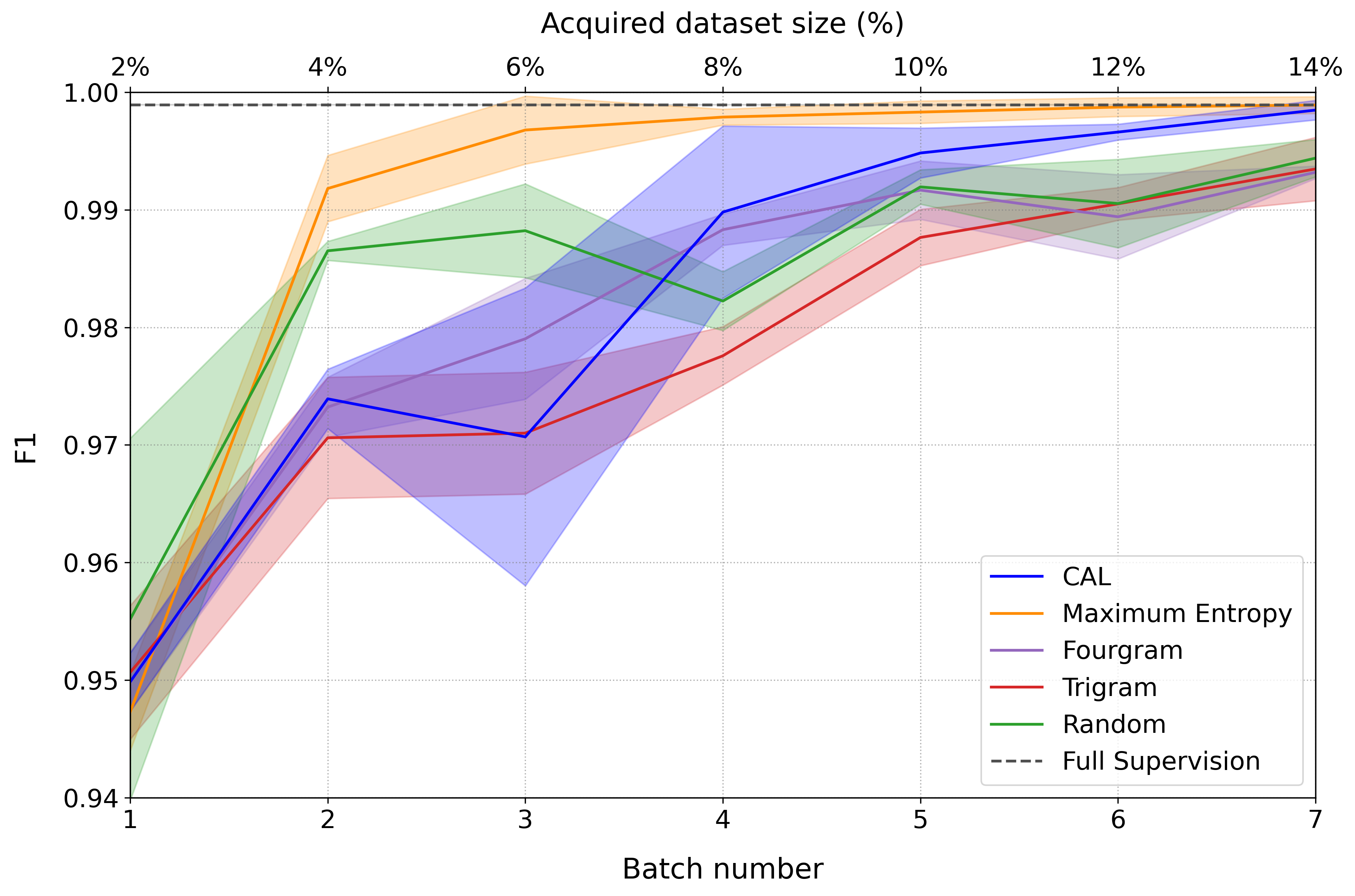}
    \caption{\textbf{Acquisition functions} (colors) are compared via macro F1 on the test set of VAST+AN (y-axis) for the model trained on $\mathcal{D}_{\text{lab}}$ associated with that AL batch (x-axis).}
  \end{subfigure}
  \hfill
  \begin{subfigure}[t]{0.4\textwidth}
    \includegraphics[width=\textwidth]{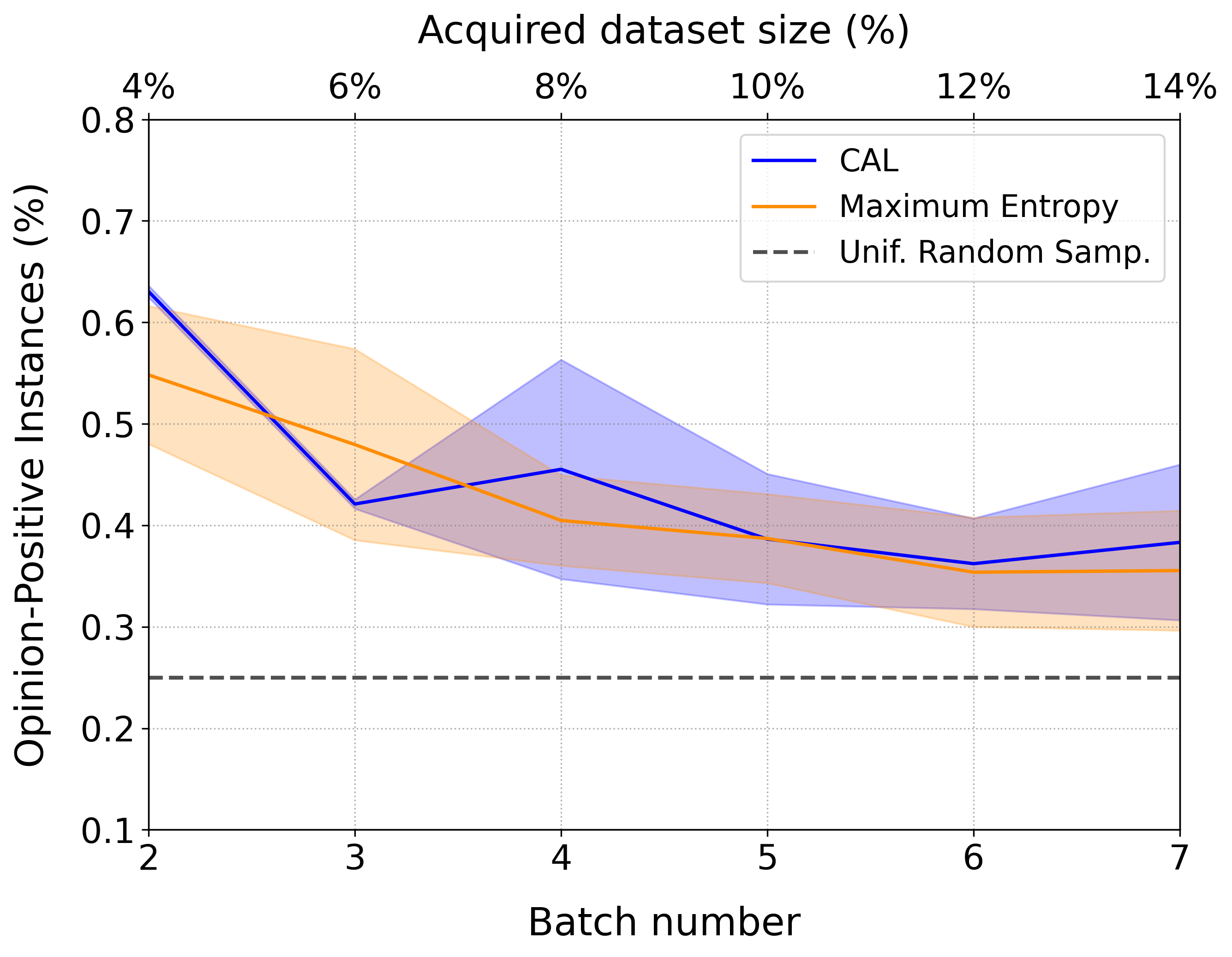}
    \caption{\textbf{Binary class distribution} of $\mathcal{D}_{\text{lab}}$ (y-axis) across active learning batches (x-axis).}
  \end{subfigure}
  \caption{\textbf{Active learning simulations} with ModernBERT and VAST+AN as $\mathcal{D}_{\text{pool}}$ with 75\% \emph{opinion-negative} and 25\% \emph{opinion-positive} labels ($n=7459$), and held-out test set ($n=3185$). 
  The mean (solid line) and standard deviation (bands) is across 5 runs with different random seeds.
  \label{fig:main-AL-results}
  }
\end{figure*}

\subsection{AL Simulations}\label{subsec:al-sims}

For AL pipelines, the ``best'' acquisition function is typically task- and model-dependent \citep{zhang2022survey}. To decide on the acquisition function for our (one-shot) deployment, we \emph{simulate} active learning by using an existing annotated dataset as a pool of unlabeled data and assign their existing labels as the ``oracle'' annotations within the AL loop, as is standard in AL \citep{margatina2023limitations}.

\textbf{Opinion Simulations.}
For our simulations, we use VAST, an annotated dataset of comments from the New York Times' ``Room for Debate'' section, where readers post short responses to policy issues or social questions \citep{allaway2020zero}.
Each comment in VAST is labeled with respect to multiple topic phrases as having a \textit{pro}, \textit{con}, or \textit{neutral} stance. 
To transform VAST into a binary classification task that matches our set-up at inference time (\S\ref{subsec:al-decisions}), we collapse any comment in VAST with at least one \textit{pro} or \textit{con} label into an \emph{opinion} label, and collapse comments with only \textit{neutral} labels into a \emph{no-opinion} label.

To match the 25\% positive and 75\% neutral class distribution observed in our prototype annotation round, we augmented VAST with additional neutral examples of randomly selected sentences from New York Times articles \citep{nyt_articles_kaggle}.  
These augmented examples are ``silver-standard'' labels because they do not have direct human verification. To estimate the false negative rate, we manually reviewed a random sample of 20 instances and found only one indicated an opinion, providing sufficient confidence in the quality for our simulations. 
We refer to the resulting binary opinion dataset as VAST+AN, where “AN” denotes the inclusion of augmented neutrals, and divide it into a train ($n=7459$) and test splits ($n=3185$). 

\textbf{Simulation Results.}
We compared the following acquisition functions: maximum entropy, CAL, n-gram density with tri-grams and four-grams, and baseline of uniform random sampling. We use an AL query batch size of 150.  
Figure~\ref{fig:main-AL-results}(a) shows that maximum entropy has both the steepest acquisition curve and reaches the level of full supervision after seeing approximately 8\% of $\mathcal{D}_{\text{pool}}$, outperforming all other methods. 
Figure~\ref{fig:main-AL-results}(b) isolates CAL and maximum entropy and shows both acquisition functions acquire a much higher percentage of stance-positive examples than the 25\% we would expect from randomly sampling $\mathcal{D}_{\text{pool}}$.
In Appendix~\ref{asec:results}, we experiment with the following: changing the model from ModernBERT to RoBERTa, changing the batch size, changing the task from classifying opinion to classifying monetary policy, and changing the class proportions to 10\%-90\%. We found our choice of acquisition function was relatively robust to these changes.   

\section{\textsc{Op-Fed} Creation}\label{sec:creation}

\textbf{Annotator training.} Prior to the annotation period, our three human annotators---all undergraduate economics majors---participated in two training sessions. The first session introduced the annotation schema and codebook. In the second session, the annotators worked together to discuss and annotate 20 example sentences drawn from the prototype rounds; see \S\ref{sec:datasheets} for additional details on the annotators. 

\textbf{AL deployment.} We then deployed our AL pipeline with maximum entropy as our acquisition function, ModernBERT as our model, an annotation budget of 1050 examples, an AL batch size of 150, and Stage 1 (opinion) as the classification task within the AL loop. On each of our seven consecutive annotation days, the three human annotators labeled the AL-selected batch of 150 instances independently and without visibility of each other's responses. Each instance was labeled for each of the five stages of our hierarchical schema. The average annotation time per instance was 50 seconds; see Table~\ref{tab:annotation_time} for additional statistics. 
For each instance and each annotation stage, we use majority vote among the three annotators to determine the final label. For opinion (Stage 1), if the majority label was \textit{ambiguous} or if there was no majority label, the instance was not used for training in the next iteration of the AL loop. 

\begin{table}[t!]
\centering
\resizebox{0.95\columnwidth}{!}{
\begin{tabular}{lrrrr}
      \toprule
      Aspect & \multicolumn{2}{c}{Unif.~Rand.} & \multicolumn{2}{c}{\textsc{Op-Fed} (w/ AL)}\\
      \toprule 
      Opinion & 24.5\% & {\small $(49/200)$} & 52.3\% &{\small $(546/1044)$}\\
      $\rightarrow$MP& 13.5\% & {\small $(27/200)$} & 26.1\% & {\small $(272/1044)$}\\
      $\rightarrow$$\rightarrow$ StanceNLI 
      & 7.5\% & {\small $(15/200)$} & 13.5\% & {\small $(141/1044)$} \\
      \bottomrule
\end{tabular}
  }
  \caption{
  \textbf{Active learning (AL) increases rates of positive labels} compared to uniform random sampling. For StanceNLI, positive labels are Entail.~$\cup$~Contr. Here, $\rightarrow$ is a reminder that the schema is \emph{hierarchical}.
  \label{tab:descriptive}}
\end{table}

\textbf{Agreement Metrics.}
In Table~\ref{tab:op-fed-final-stats}, we provide descriptive statistics of the final dataset---after filtering to instances that received a majority vote---and we report chance-adjusted agreement scores of Fleiss' Kappa ($\kappa$) \citep{fleiss2003statistical} and Kripendorff Alpha ($\alpha$) \citep{krippendorff2004content}.
\textsc{Op-Fed} achieved moderate agreement levels for Stage 1 opinion ($\kappa=0.57$), Stage 2 MP ($\alpha=0.62$), Stage 3 MP context ($\alpha=0.72$), Stage 4 Stance NLI ($\alpha=0.55$), and Stage 5 Stance NLI Context ($\alpha=0.67)$.  While these agreement rates may seem low compared to simple NLP tasks, they are on par with other complex tasks in computational social science, e.g., classifying ``grand'' versus ``formal'' legal reasoning in U.S.~Supreme Court opinions which reported $\alpha=0.63$ \citep{thalken2023modeling}.

For Stages 3 and 5, most of instances required at least utterance-level context (Figure~\ref{fig:context}), confirming \textbf{Challenge I}.  
Table~\ref{tab:fleiss_kappa} breaks down the Fleiss Kappa scores for Stage 1 by AL batch and annotator pair. The agreement scores decrease with each AL batch. We hypothesize this is due to the most uncertain examples---selected by the maximum entropy acquisition function---also being the more challenging examples for human annotators.  

\textbf{AL analysis.} Our AL pipeline successfully addressed our goal of mitigating the class imbalance (\textbf{Challenge II}). In Table~\ref{tab:descriptive}, we find the number of positive instances roughly doubled across the three target aspects when comparing our prototype annotation round---sampled uniformly at random---to \textsc{Op-Fed}---sampled via AL.

\begin{figure}[t]
  \centering
   \includegraphics[width=0.95\columnwidth]{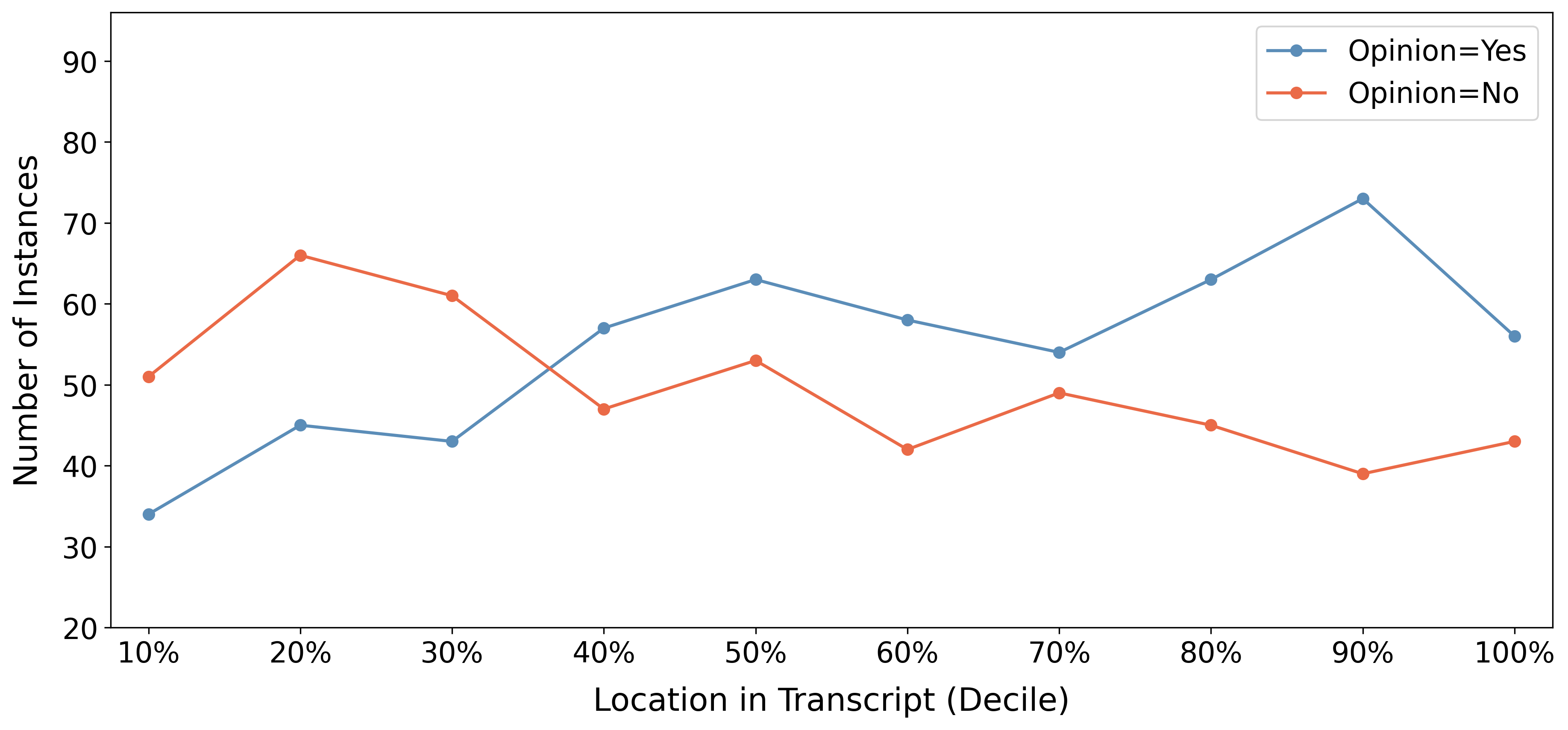}
   \caption{
   \textbf{Opinions are expressed later in transcripts.} Distribution of \textsc{Op-Fed} gold-standard opinion labels by their location in FOMC transcripts.}
   \label{fig:face_validity_1}
\end{figure}

\textbf{Face Validity.}
To assess the \emph{face validity}---whether measurements ``look plausible'' in the social sciences \citep{jacobs2021measurement}---of the opinion labels, Figure~\ref{fig:face_validity_1} shows that opinion-positive statements are more likely to occur later in the transcripts, with the highest being in the 80-90\% decile. 
This pattern is consistent with the known structure of FOMC meetings---in the latter portion of the meeting, participants engage in the ``policy go-around,'' in which each member outlines their preferred policy stance---and lends face validity to our sampling and annotation efforts.

\section{Baseline Experiments}\label{sec:experiments} 

\begin{table*}[t]
\centering
\resizebox{0.98\linewidth}{!}{
\begin{tabular}{l l l r r r rr}
\toprule
& \textbf{Model} & \textbf{Checkpoint} & \textbf{1-Opinion} & \textbf{2-MP} & \textbf{3-MP Ctxt.} & \textbf{4-StanceNLI}  & \textbf{5-Stance Ctxt.}\\
& & & {\small($n=1042$)} & {\small($n=485$)} &  {\small($n=242$)} & {\small($n=159$)} & {\small($n=162$)}\\
\midrule
\multirow{4}{*}{Baselines} & Majority Class & -- & 0.52&0.56&0.59&0.59&0.85 \\ 
& BOW+LogReg & -- & 0.61&0.60&0.80&0.31&0.73\\
& Human min.$^*$ & -- & 
0.85&0.85&0.86&0.79&0.85
\\
& Human max.$^*$ & -- & 
0.95&0.89&0.89&0.89&0.93 \\
\hline
\multirow{4}{*}{Zero-Shot}
& GPT-5 & \texttt{gpt-5-2025-08-07} & 0.72&0.74&0.28&0.47&0.28\\ 
& GPT-5 nano & \texttt{gpt-5-nano-2025-08-27} & 0.75&0.73&0.33&0.46&0.28\\
& Claude Opus 4.1 & \texttt{claude-opus-4-1-20250805} &
0.80 &0.74&0.24&0.61&0.29\\
& DeepSeek-V3.1 & \texttt{deepseek-chat-2025-08-21} & 0.75&0.76&0.47&0.29&0.15\\
\bottomrule
\end{tabular}
}
\caption{\textbf{Accuracy of baseline models.} 
Stages 3 and 5 are each a binary classification task for whether context is needed beyond the sentence. Here, $n$ is the number of examples after removing instances with an \emph{ambiguous} label.
Corresponding F1 scores are in Table~\ref{tab:f1_results} and prompts in \S\ref{appendix:prompts}.
\label{tab:stance_baselines}
}
\end{table*}

\subsection{Zero-Shot LLM Performance}

LLMs have achieved relatively high performance for zero-shot classification tasks, i.e. performing an unseen task with no supervised examples \citep[\textit{inter alia}]{brown2020language,wei2021finetuned,sanh2021multitask}. \citet{peskoff2023gpt} use GPT-4 zero-shot on FOMC transcripts with the assumption that the zero-shot accuracy is sufficient. 
With the creation of \textsc{Op-Fed}, we can answer one of the motivating questions for this work: what is the zero-shot performance of LLMs for the FOMC tasks we have described with our schema?  

In our experimental set-up, we subset to \textsc{Op-Fed} instances with non-ambiguous labels.\footnote{This makes Task 4 a multi-class (with three classes) and the other tasks binary. For Tasks 2 and 5, we provide the models with maximum context for the target sentence: the utterance of target sentence and the 200 previous tokens to that utterance.}  
We compare Open AI's \emph{GPT-5} and \emph{GPT-5-nano} models, Anthropic's \emph{Claude Opus 4.1}, and \emph{DeepSeek-V3.1}. We use the recommended structured output format for each model; see \S\ref{appendix:prompts} for all prompting details. For comparison, we also implement baselines of predicting the majority class; four-fold cross validation accuracy with a bag-of-words logistic regression model (see \S\ref{app:bow}); and estimates of the minimum and maximum human baselines, i.e.~the accuracy achieved by one (out of three) annotators when compared against the aggregated gold-standard.

Examining Table~\ref{tab:stance_baselines}, LLMs zero-shot accuracy is relatively high on Task 1, $0.72-0.8$, and Task 2, $0.73-0.76$, but still slightly below the maximum human baselines on the two tasks with $0.95$ and $0.89$ respectively.  For Task 4, accuracy is relatively low across all models with the highest accuracy ($0.61$) from \emph{Claude Opus 4.1}, which only slightly outperforms predicting the majority class. Notably, \emph{DeepSeek-V3.1} on Task 4 has a lower accuracy than our bag-of-words logistic regression baseline, $0.29$ versus $0.31$ respectively. All LLMs perform poorly on predicting whether more context is needed (Tasks 3 and 5) with lower accuracy than both the  majority class and bag-of-words logistic regression baselines. Thus, our recommendation to economists would be to perform additional fine-tuning of LLMs on gold-standard data if their goal is Task 4; see \citet{halterman2025codebook}. 

\subsection{Confidence Score Calibration.} 
Downstream FOMC analyses likely also need \emph{confidence scores} on individual instances in addition to class predictions; see \citet{angelopoulos2023prediction}. We prompt the zero-shot LLMs in our experiments to generate ``verbalized confidence scores''---values between 0.0 and 1.0 expressed in token-space---since \citet{tian2023just} found these are often better calibrated than model conditional probabilities.
We evaluate with \textbf{expected calibration error} (ECE) which partitions the confidence score into $M$ equally sized bins \citep{guo2017calibration}.\footnote{We use $M=20$. We leave to future work alternative calibration estimation methods \citep{nguyen2015posterior}.}
In Table~\ref{tab:calibration_error}, we report ECE for Tasks 1 and 2 (those with decent zero-shot accuracy), and we find the zero-shot models are moderate to poorly calibrated; \emph{Clause Opus 4.1} achieves the lowest calibration error on Task 1 (ECE=$0.09$) and \emph{GPT-5 nano} achieves the lowest calibration error on Task 2 (ECE=$0.06$), but error can be substantial, e.g., ECE=$0.20$ on Task 1 for \emph{GPT-5}. 
For context, tis is much lower than \citet{tian2023just}'s findings that \texttt{gpt-3.5-turbo}'s verbalized confidence scores can achieve ECE as low as 0.05 on standard NLP benchmarks. We see improving confidence score calibration as an area of future work where our dataset could be useful, e.g.,~Platt scaling \citep{platt1999probabilistic} with our gold-standard labels.

%evaluate_calibration.py
\begin{table}[t]
\centering
\resizebox{0.65\columnwidth}{!}{
\begin{tabular}{lrr}
\toprule
\textbf{Model} & \textbf{1-Opinion} & \textbf{2-MP}\\
\toprule
GPT-5 & 
0.20&0.14
\\ 
GPT-5 nano & 
0.13&0.06
\\
Claude Opus 4.1 &
0.09&0.17
\\
DeepSeek-V3.1 & 
0.13&0.13\\
\bottomrule
\end{tabular}
}
\caption{\textbf{Expected calibration error (ECE)} of each LLM's verbalized (generated) confidence scores. 
\label{tab:calibration_error}
}
\end{table}

\section{Conclusion and Future Work}\label{sec:future}

In this work, we release \textsc{Op-Fed}, a human-annotated dataset of FOMC transcripts with hierarchical labels for opinion, monetary policy, stance toward monetary policy as well as the context needed beyond the sentence-level.
Using active learning substantially increased the yields of positive instances in our dataset. 
Future work could likely improve LLM accuracy and confidence score calibration by shifting to supervised fine-tuning and/or other iterative inference techniques. 
Other work could use our hierarchical schema and \textsc{Op-Fed} as the seed dataset in a new AL loop to create an even larger dataset of instances. Finally, economists could combine our gold-standard annotations with zero-shot LLM predictions to improve economic analyses, i.e.~\citet{angelopoulos2023prediction}'s method. 

\clearpage
\newpage

\section*{Acknowledgments}

We thank Yanni Kakouris and Aidan Casey for their work on the annotations. Thanks to Mark Hopkins for feedback on an early draft, to Kenneth Kuttner for a conversation about the FOMC, and to anonymous ARR reviewers for helpful suggestions. 
KK acknowledges YI grant from AI2.

\section*{Limitations}

Given that these transcripts are publicly available government data, we see very few potential risks with this project or dataset. 
We identify FOMC speakers by name, but these speakers are public figures whose consent is implicit given their professional roles. 

There are limitations to our work due to our annotation budget, using undergraduate annotators, and benchmarking with proprietary, closed-weight models. First, given our fixed annotation budget, we successfully increased the class proportions by using active learning; however, we believe this could be improved by future work with larger annotation budgets. 

Second, we used three ``in-house,'' trained annotators who were undergraduate students majoring in economics. We hypothesized undergraduates would have higher annotation quality compared to crowd-workers. However, we also hypothesize that domain-experts---who have greater expertise with the Federal Reserve and monetary policy---may annotate instances with higher quality. 

Our final limitation is our use of proprietary, closed-weight LLMs on \textsc{Op-Fed}. As \citet{palmer2024using} advise, ``using proprietary language models in academic research requires explicit justification.'' Proprietary models lack details about their training data, which limits transparency and reproducibility. However, we use these models as baselines because they are likely to be the first choice of applied researchers (i.e.,~economists) so we believe it is important to benchmark their accuracy. 

\bibliography{bib,custom}

\appendix

\section{Artifacts: Citations and Licenses}

See Table~\ref{tab:artifacts} for artifact citations, versions, and licenses. The use of all artifacts is consistent with their intended use. 

\begin{table*}[t!]
\centering
\resizebox{0.98\linewidth}{!}{
\begin{tabular}{lll
>{\raggedright\arraybackslash}
p{5cm}}
\toprule
\textbf{Artifact} & \textbf{Version (date)} & \textbf{License} & \textbf{Citation} \\ \toprule
spaCy & 3.8.7 & MIT & \citet{honnibal-johnson-2017-spacy} \\
\hline
ConvoKit & 3.4.1 & MIT & \citet{chang2020convokit} \\ \hline
ModernBERT & ModernBERT-base (2024-12-19) & Apache 2.0 & \citet{warner2024modernbert} \\ \hline
RoBERTa & roberta-base (v1, 2019) & MIT & \citet{liu2019roberta} \\ \hline
GPT-5 & \texttt{gpt-5-2025-08-07} & Proprietary (API) & No publication \\ \hline 
GPT-5-nano & \texttt{gpt-5-nano-2025-08-27 } & Proprietary (API) & No publication \\ \hline
Claude Opus 4.1 & \texttt{claude-opus-4-1-20250805} & Proprietary (API) & No publication\\ \hline
DeepSeek-V3.1 &  \texttt{deepseek-chat-2025-08-21} & Proprietary (API) & No publication\\
\bottomrule
\end{tabular}
}
\caption{Latest version, citation, and license for NLP artifacts used. \label{tab:artifacts}}
\end{table*}

\section{Datasheets for Datasets}\label{sec:datasheets}

In this section, we report our answers to Datasheets for Datasets \citep{gebru2021datasheets}. 

\subsection{Motivation for Dataset Creation}

\noindent\textbf{Why was the dataset created?} \\
This dataset was created to analyzes the opinions and stances U.S. Federal Open Market Committee (FOMC) members in their verbatim meeting transcripts. Existing datasets on FOMC communications either lack speaker attribution, rely on simple dictionary-based or bag-of-words approaches, or are limited to curated sources such as press releases and meeting minutes. \textsc{Op-Fed} addresses these limitations by annotating transcripts with a novel hierarchical annotation schema that captures opinion, monetary policy references, and stance directionality---while preserving speaker identity and contextual nuance.

\vspace{0.5em}

\noindent\textbf{What (other) tasks could the dataset be used for?} \\
This dataset could be used to support research on group decision-making, hierarchical multi-label classification, and opinion classification and stance detection in complex policy settings. 

\vspace{0.5em}

\noindent\textbf{Has the dataset been used for any tasks already?} \\
At the time of publication, only the original paper. 

\vspace{0.5em}

\noindent\textbf{Who funded the creation of the dataset?} \\
Funding was provided via the senior author's start-up funds from their institution. 

\subsection{Dataset Composition}

\noindent\textbf{What are the instances?} \\
Instances are individual sentences extracted from raw FOMC meeting transcripts. Each instance is annotated using a five-stage hierarchical schema that captures whether the speaker expresses (1) an opinion, (2) whether that opinion refers to monetary policy, (3) how much context (beyond the sentence-level) was needed to determine the monetary policy reference, (4) whether the sentence entails, contradicts, or is neutral toward the hypothesis “I think we should tighten monetary policy,” and (5) how much context (beyond the sentence-level) was needed to assign that inference label.

\vspace{0.5em}

\noindent\textbf{Are relationships between instances made explicit in the data?}\\
Not explicitly; while context (e.g., preceding sentences) is tracked, no graph or linkage structure is included.

\vspace{0.5em}

\noindent\textbf{How many instances of each type are there?} \\
There are 1044 instances in \textsc{Op-Fed} and 1050 instances in the unaggregated dataset (that is, the one that releases the labels for each of the three annotators). 

\vspace{0.5em}

\noindent\textbf{What data does each instance consist of?} \\
Each instance consists of a target sentence, associated context (including the full utterance, the five preceding sentences, and up to 200+ preceding tokens---rounded to full sentences), and labels from the five-stage hierarchical schema. Annotations are provided both as aggregate (majority vote) labels and as raw labels from each of the three annotators.

\vspace{0.5em}

\noindent\textbf{Is everything included or does the data rely on external resources?} \\
Everything is included.

\vspace{0.5em}

\noindent\textbf{Are there recommended data splits or evaluation measures?} \\
No official splits are provided. 

\vspace{0.5em}

\noindent\textbf{What experiments were initially run on this dataset?} \\
We experiment with zero-shot classification using closed-weight proprietary LLMs. 

\subsection{Data Collection Process}

\noindent\textbf{How was the data collected?} \\ 
Raw transcripts were sourced from the ConvoKit version of the FOMC corpus. Sentences were sampled for annotation using active learning.

\vspace{0.5em}

\noindent\textbf{Who was involved in the data collection process?} \\
The dataset was annotated by three undergraduate economics majors from a college in the United States, two of whom were hired for the project and the third being an author on this paper. The hired annotators were paid \$30/hour with total compensation amounting to \$420 per annotator.

\vspace{0.5em}

\noindent\textbf{Over what time-frame was the data collected?}
Annotation occurred over seven consecutive days, from March 31 to April 6, 2025, following annotator training.

\vspace{0.5em}

\noindent\textbf{How was the data associated with each instance acquired?} \\
The data was acquired via human annotation in batches of 150 instances, using a codebook to guide labeling decisions.

\vspace{0.5em}

\noindent\textbf{Does the dataset contain all possible instances?} \\
No. It is a sample from a larger set of 280,975 sentences.

\vspace{0.5em}

\noindent\textbf{If the dataset is a sample, then what is the population?} \\
All sentences in the FOMC transcripts between 1977 and 2008, after preprocessing.

\vspace{0.5em}

\noindent\textbf{Is there information missing from the dataset and why?} \\
No data is missing.

\vspace{0.5em}

\noindent\textbf{Are there any known errors, sources of noise, or redundancies in the data?} \\
No.

\subsection{Data Preprocessing}

\noindent\textbf{What preprocessing/cleaning was done?} \\
Utterances with less than 4 or more than 500 tokens were excluded from the unlabeled pool. We used sentence segmentation to split sentences.

\vspace{0.5em}

\noindent\textbf{Was the “raw” data saved in addition to the preprocessed/cleaned data?} \\
Yes, the raw data exists from ConvoKit. 

\vspace{0.5em}

\noindent\textbf{Is the preprocessing software available?} \\
No.

\vspace{0.5em}

\noindent\textbf{Does this dataset collection/processing procedure achieve the motivation for creating the dataset stated in the first section of this datasheet?} \\
Yes.

\subsection{Dataset Distribution}

\noindent\textbf{How is the dataset distributed?} \\
At the time of publication, the dataset will be distributed via the authors' GitHub and will also be posted to HuggingFace. 

\vspace{0.5em}

\noindent\textbf{When will the dataset be released/first distributed?} \\
At the time of publication. 

\vspace{0.5em}

\noindent\textbf{What license (if any) is it distributed under?} \\
We will release a license at the time of publication. 

\vspace{0.5em}

\noindent\textbf{Are there any fees or access/export restrictions?} \\
No.

\subsection{Dataset Maintenance}

\noindent\textbf{Who is supporting/hosting/maintaining the dataset?} \\
% Katie Keith.
The last author on this paper. 

\vspace{0.5em}

\noindent\textbf{Will the dataset be updated?} \\
No updates are planned.

\vspace{0.5em}

\noindent\textbf{If the dataset becomes obsolete how will this be communicated?} \\
This will be posted on the project's GitHub page. 

\vspace{0.5em}

\noindent\textbf{Is there a repository to link to any/all papers/systems that use this dataset?} \\
No.

\vspace{0.5em}

\noindent\textbf{If others want to extend/augment/build on this dataset, is there a mechanism for them to do so?} \\
They may download and build upon it freely.

\subsection{Legal \& Ethical Considerations}

\noindent\textbf{If the dataset relates to people (e.g., their attributes) or was generated by people, were they informed about the data collection?} \\
Not applicable. The data are public-domain government transcripts.

\vspace{0.5em}

\noindent\textbf{If it relates to other ethically protected subjects, have appropriate obligations been met?} \\
Not applicable. 

\vspace{0.5em}

\noindent\textbf{If it relates to people, were there any ethical review applications/reviews/approvals?} \\
No.

\vspace{0.5em}

\noindent\textbf{If it relates to people, were they told what the dataset would be used for and did they consent? What community norms exist for data collected from human communications?} \\
Not applicable. 

\vspace{0.5em}

\noindent\textbf{If it relates to people, could this dataset expose people to harm or legal action?} \\
Not applicable. 

\vspace{0.5em}

\noindent\textbf{If it relates to people, does it unfairly advantage or disadvantage a particular social group?} \\
Not applicable. 

\vspace{0.5em}

\noindent\textbf{If it relates to people, were they provided with privacy guarantees?} \\
Not applicable. 

\vspace{0.5em}

\noindent\textbf{Does the dataset comply with the EU General Data Protection Regulation (GDPR)?} \\
Not applicable. The data are public-domain government transcripts.

\vspace{0.5em}

\noindent\textbf{Does the dataset contain information that might be considered sensitive or confidential?} \\
No.

\vspace{0.5em}

\noindent\textbf{Does the dataset contain information that might be considered inappropriate or offensive?} \\
No.

\section{Additional Analysis of \textsc{Op-Fed}}

%code/descriptive/descriptive.py
\begin{table}[t!]
  \centering
  \resizebox{0.95\linewidth}{!}{ %makes it fit within the margin limits
      \begin{tabular}{llr}
      \toprule
      \multicolumn{2}{l}{\textbf{Unlabeled Pool}, $\mathcal{D}_{\text{pool}}$} & $n$ \\
      \toprule
      Num.~Transcripts & & 286\\
      Num.~Sentences & & 280,975\\
      \toprule
      \multicolumn{2}{l}{\textbf{\textsc{Op-Fed}}, Final $\mathcal{D}_{\text{lab}}$} & $n$ \\
      \toprule
      1-Opinion & Yes & 546\\
      ($\kappa=0.57$) & No & 496\\ 
      & Ambiguous & 2\\ 
      \hline
      2-MP & Yes & 272\\
      ($\alpha=0.62$) & No & 214\\
      & Ambiguous & 1\\
      \hline
      3-MP Context & Sentence & 99\\
      ($\alpha=0.72$) & Utterance & 119\\
      & -5 sents. & 20\\ 
      & -200+ toks. & 4\\
      \hline
      4- StanceNLI & Entailment & 47\\ 
      ($\alpha=0.55$) & Neutral & 18\\
      & Contradiction & 94\\ 
      & Ambiguous & 15\\ 
      \hline
      5-StanceNLI Context & Sentence & 24\\
      ($\alpha=0.67$) & Utterance & 111\\
      & -5 sents. & 22\\ 
      & -200+ toks. & 5\\ 
      \bottomrule
  \end{tabular}
  }
  \caption{
  \textbf{Descriptive statistics} of \textsc{Op-Fed}---after taking the majority vote from the three human annotators---including Fleiss' kappa ($\kappa$) and Kripendorff's alpha ($\alpha$) for this subset.
  \label{tab:op-fed-final-stats}}
\end{table}

\begin{table}[t]
\renewcommand{\arraystretch}{1.2}
\centering
\resizebox{0.98\linewidth}{!}{
\begin{tabular}{
>{\raggedright\arraybackslash}p{1cm}
>{\centering\arraybackslash}p{2cm}
>{\centering\arraybackslash}p{2cm}
>{\centering\arraybackslash}p{2cm}
>{\centering\arraybackslash}p{2cm}}
\toprule
\textbf{} 
& \multicolumn{4}{c}{\textbf{Annotators}} \\
\textbf{Batch} & All & 1 and 2 & 2 and 3 & 1 and 3 \\
\midrule
1     & \cellcolor{green!20}0.647 & \cellcolor{green!20}0.663 & \cellcolor{yellow!20}0.558 & \cellcolor{green!20}0.716 \\
2     & \cellcolor{yellow!20}0.572 & \cellcolor{orange!20}0.499 & \cellcolor{orange!20}0.420 & \cellcolor{green!40}0.808 \\
3     & \cellcolor{green!20}0.681 & \cellcolor{yellow!20}0.639 & \cellcolor{green!20}0.684 & \cellcolor{green!20}0.715 \\
4     & \cellcolor{yellow!20}0.541 & \cellcolor{yellow!20}0.539 & \cellcolor{yellow!20}0.492 & \cellcolor{green!20}0.589 \\
5     & \cellcolor{yellow!20}0.512 & \cellcolor{yellow!20}0.584 & \cellcolor{orange!20}0.413 & \cellcolor{yellow!20}0.538 \\
6     & \cellcolor{orange!20}0.457 & \cellcolor{yellow!20}0.620 & \cellcolor{red!20}0.308 & \cellcolor{orange!20}0.433 \\
7     & \cellcolor{orange!20}0.411 & \cellcolor{orange!20}0.430 & \cellcolor{red!20}0.202 & \cellcolor{yellow!20}0.609 \\
\midrule
\textbf{Total} & \cellcolor{yellow!20}0.567 & \cellcolor{yellow!20}0.586 & \cellcolor{yellow!20}0.463 & \cellcolor{green!20}0.649 \\
\bottomrule
\end{tabular}
}
\caption{
\textbf{Fleiss' Kappa scores for opinion} (Stage 1) by AL batch and annotator pair.}
\label{tab:fleiss_kappa}
\end{table}

\begin{figure}[h]
\centering
\includegraphics[width=0.5\textwidth]{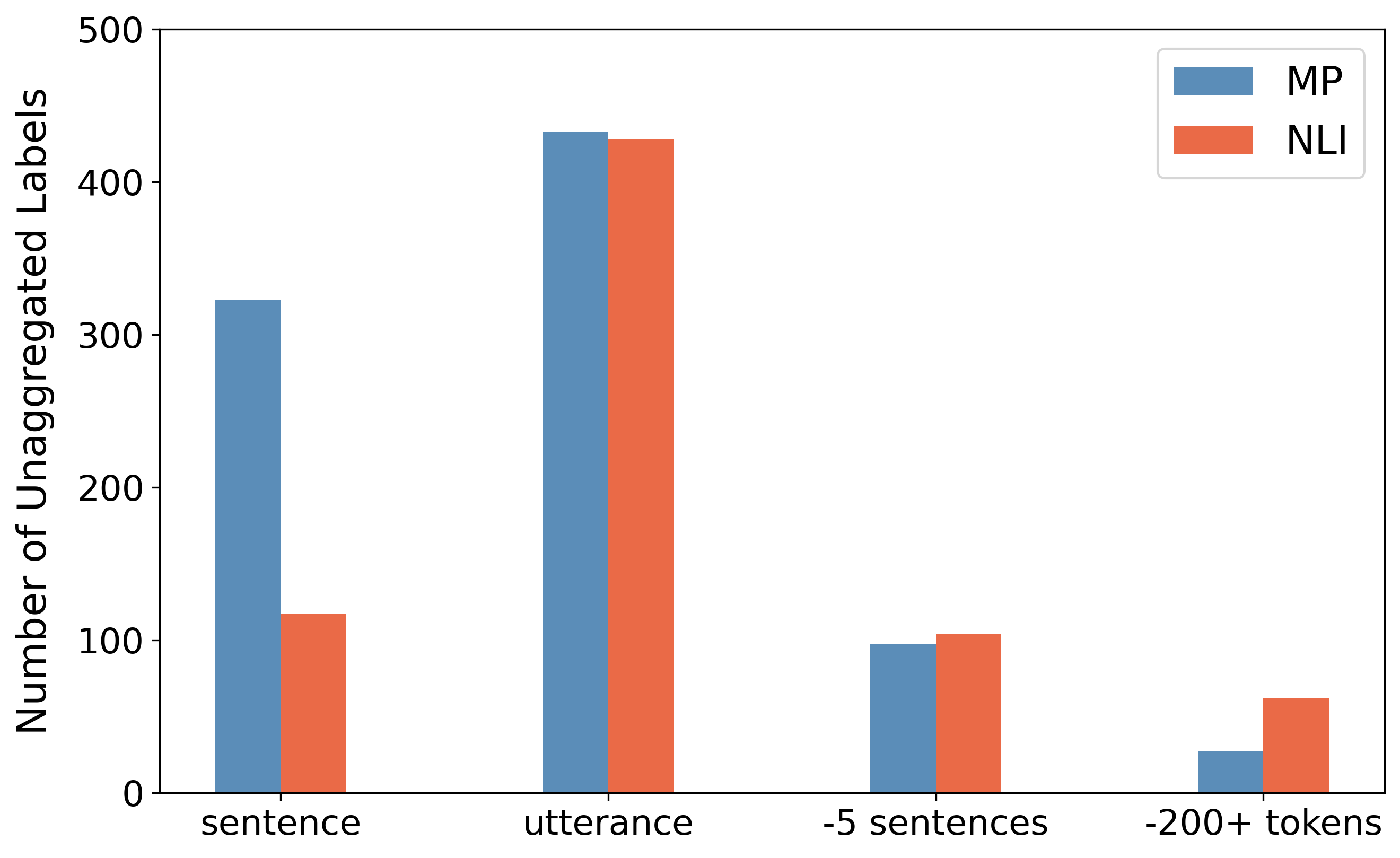}
\caption{Context label distribution across the three
annotators for Stage 3 - MP Context and Stage 5 - Stance NLI Context. \label{fig:context}}
\end{figure}

\begin{table*}[h]
\renewcommand{\arraystretch}{1.2}
\centering
\begin{tabular}{
  >{\raggedright\arraybackslash}p{1cm}  % Left-align first column
   >{\centering\arraybackslash}p{1.5cm}
    >{\centering\arraybackslash}p{1.5cm}
    >{\centering\arraybackslash}p{1.5cm}}
\toprule
& \multicolumn{3}{c}{\textbf{Annotator}} \\
\cline{2-4}
\textbf{Batch} & 1 & 2 & 3 \\
\hline
1 &  90  & 120 & 145 \\
2 & 135  & 105 & 180 \\
3 &  90  & 135 & 105 \\
4 & 120  & 120 & 120 \\
5 & 130  & 140 & 130 \\
6 & 120  & 135 & 135 \\
7 & 130  & 120 & 120 \\
\hline
\textbf{Total} & 815 & 875 & 935 \\
\bottomrule
\end{tabular}
\caption{Time to complete annotations (in minutes) by annotator and batch.}
\label{tab:annotation_time}
\end{table*}

%code/descriptive/hand_selected_examples.ipynb
\begin{table*}[t]
\renewcommand{\arraystretch}{1.3}
\centering
\resizebox{0.7\linewidth}{!}{
\begin{tabular}{
        l 
        >{\raggedright\arraybackslash}m{0.17\textwidth}
        >{\raggedright\arraybackslash}m{0.8\textwidth}}
  \toprule
  \toprule
      \textbf{Unique id} & \textbf{Sentence} & \textbf{MP Context or StanceNLI Context} \\
     \toprule
     \toprule
     19811222\_230\_4 &
     \textit{``But the discrepancies in the numbers were even larger than we had.''} & n/a 
     \\
     \hline 
     9910206\_54\_6 &
     \textit{``If the Congress feels comfortable with \hl{the arrangement}, I'm certainly on board in this context.''} & n/a \\
     \hline
     19810203\_410\_1 &
     \textit{``I don't know how important \hl{all of this} is.''} & 
     \textbf{-5 sentences:}
     'speaker': 'MR. GRAMLEY.', 'text': "It seems to me, however, that the market is going to interpret your suggestion in a very different way. If the fed funds rate on the day of the Committee meeting is 17.42 percent and we say 300 basis points on either side, the newspapers would say that they have figured out that the Fed's range is 14.42 to 20.42 percent. It makes a lot more sense to say it's 14 to 20 percent. I think that's what they're going to do. Then if we start playing games and say we're going to narrow the range down to 200 basis points on either side because we don't want rates to go up too far, they're surely going to say we are playing the rates instead of the money supply."
     \\
     \hline
     19791006\_304\_14 &
     \textit{``\hl{That} sounds like a good idea.''}
     & 
     \textbf{Utterance:}
     A 1 percentage point increase. Under those circumstances, let me modify my proposal slightly. We work with a 4 . 6 or 4.5 percent target for M1, recognizing that people would rather see M1 growth come in somewhat lower, certainly, than higher. But that to us is a satisfactory target. In an uncertain world, all other things equal, the money market nice and equitable, expectations changing nicely, things settling down, it's possible we would feel better if it came in below than if we were embarrassed by it being too high. We have a discount rate of 12 percent. We have the reserve requirement change, which I think will be 8 percent, on a basket of managed liabilities. That is about equivalent to an added cost of 1 percent on those [liabilities]. That's the effect that has, mechanically. I don't know what that does to the prime rate, but [the added cost to banks] is just marginal. If the banks want to be mean--I've got to speak to the MA--they raise the prime rate and say that's our marginal cost of funds. If they want to be reasonable, they don't because that cost is only going to apply to a very small amount at the margin. I would be inclined to tell them at the ABA that they shouldn't reach forever on the interest rate. That sounds like a good idea. In and of itself, I think [all] that means some increase in the federal funds rate. How much of an increase? Although this scenario only puts the discount rate at about or slightly above the current federal funds rate, it nonetheless, all other things equal, puts some upward pressure on the federal funds rate. In terms of the range, Phil suggested 11-1/2 to 14-1/2 percent. As I told you, it makes me nervous to think of [the rate] going up and getting locked in at a higher level. But if you want to put the upper end higher, I reserve the right to consult if the funds rate were to begin getting up that high. We could have that understanding. I am not talking about for a day; I think it would be fine if it went there for a day or a bit beyond.
     \\
    \hline
     19850213\_474\_1 &
     \textit{``I would go with \hl{"B"} also.''} &
     \textbf{-5 sentences:}
     'speaker': 'MR. MARTIN.', 'text': 'It seems to me that the odds are that M1 will exceed 8 percent. My only discomfort is that 8-1/4 percent may not be the correct fed funds rate for the short run; maybe it should be 8-1/2 percent. But I take it that there is enough flexibility built into our procedures [to allow for] an 8-1/2 percent rate if we need it to get 8 percent [M1 growth] or whatever that is--less than the 10 percent and 12 percent that we\'ve been experiencing in recent months. Given that assumption, I would go along with "B."', 'speaker': 'CHAIRMAN VOLCKER.', 'text': 'Mr. Boykin.'
     \\
     \hline
        19930707\_543\_1
     &
     \textit{``Well, I'm clearly not in tune with the other members of the FOMC.''} & 
     \textbf{Utterance:}
     Well, I'm clearly not in tune with the other members of the FOMC. I don't see any need to wait for any information. The markets provide indications every day as to whether or not we've provided more liquidity than is called for. And when we lowered the fed funds rate from 4 percent to 3 percent, my guess is that made very, very little difference in the rate of real GDP growth. The destabilizing factors that have led people to hold, shift, or use their balances as much as they have was, I think, a drag against the stimulus that was already in place. A 4 percent fed funds rate already was providing a lot of stimulus. We already had pegged the fed funds rate well below the natural rate of interest. We now have evidence, and we see it in our own staff's forecast, that the expected rate of inflation has moved up 0.6 or 0.7 percent just [since the May FOMC] meeting. Now, in that environment, policy is not stable. We do not have the same policy that we had at the last meeting because the real rate of interest by our best estimate has fallen to even more sharply negative territory. We clearly see in the price of gold that people are making bets out there. And we continue to lock in to a fed funds rate that will only aggravate that kind of speculation and it only detracts from the role of the U.S. currency as the stable currency for the world. The cost to the world of the United States pursuing inflationary policies in the late 1960s and the 1970s is unbelievable. We're still paying the cost because many other central banks with no confidence in us think they have to be Rambo-like in beating their chests because in some sense they've got a track for the inflation-induced environment that the Federal Reserve as the world's reserve currency provides. This is the time for us to move real interest rates back at least to zero. I would be satisfied to do a measly 1/4 percentage point, which would not get us back to zero, but there would be intervention value in that. There wouldn't be any harm in it. Is there anyone who really believes the U.S. economy, regardless of what is done on the fiscal [side], is going to suffer because the funds rate goes up from 3 to 3-1/4 percent? Now, I read Henry Kauffman, as many of you must have, and it's just absurd. Well, I must be the one that's absurd! Thank you.\\
      \bottomrule
      \bottomrule
    \end{tabular}

}
  \caption{
  This table provides the context for Table~\ref{tab:example_annotations} and adds the maximum context needed for either the MP or Stance NLI label. 
  \label{tab:example_annotations-context}}
\end{table*}

\begin{table*}[t]
\renewcommand{\arraystretch}{1.2}
\centering
\resizebox{0.98\linewidth}{!}{
\begin{tabular}{l l l r r r rr}
\toprule
& \textbf{Model} & \textbf{Checkpoint} & \textbf{1-Opinion} & \textbf{2-MP} & \textbf{3-MP Ctxt.} & \textbf{4-StanceNLI}  & \textbf{5-Stance Ctxt.}\\
& & & {\small($n=1042$)} & {\small($n=485$)} &  {\small($n=242$)} & {\small($n=159$)} & {\small($n=162$)}\\
\midrule
\multirow{4}{*}{Baselines} & Majority Class & -- & 0.52&0.56&0.59&0.44&0.85\\
 & BOW+LogReg & -- & 
0.61&0.60&0.80&0.31&0.75 \\
& Human min.$*$ & -- &
0.85&0.85&0.86&0.80&0.85 \\
& Human max$^*$ & -- & 
0.95&0.89&0.89&0.89&0.93\\
\hline
\multirow{4}{*}{Zero-Shot}
& GPT-5 & \texttt{gpt-5-2025-08-07} &
0.72&0.74&0.28&0.52&0.28\\ 
& GPT-5 nano & \texttt{gpt-5-nano-2025-08-27} & 0.75&0.73&0.33&0.48&0.28\\
& Claude Opus 4.1 & \texttt{claude-opus-4-1-20250805} &
0.80&0.74&0.24&0.63&0.29\\
& DeepSeek-V3.1 & \texttt{deepseek-chat-2025-08-21} & 0.75&0.76&0.47&0.31&0.15\\
\bottomrule
\end{tabular}
}
\caption{\textbf{F1 scores of baseline models.} These are for comparison to the accuracy scores from Table~\ref{tab:stance_baselines}. Task 4 is a weighted F1 score, all the rest are (micro) F1 scores for a binary classification task. 
\label{tab:f1_results}
}
\end{table*}

In this section, we provide additional analyses of \textsc{Op-Fed}. 
Table~\ref{tab:op-fed-final-stats} provides the label distribution breakdown of the dataset and inner-annotator statistics of the dataset. These inner-annotator agreement rates are for the subset reported in the table (i.e., after discarding instances that did not achieve a 2/3 majority vote by annotators). 

Table~\ref{tab:fleiss_kappa} shows the Fleiss Kappa scores for stage 1 (opinion) by AL batch number and by annotator pairs. Table~\ref{tab:annotation_time} reports the annotation time per annotator. 

\section{Additional Modeling and Compute Details}

All experiments are run on a local server that has the following NVIDIA GPUs: 16 $\times$ A6000s and 4 $\times$ A100s. See Table~\ref{tab:hyperparams} for hyperparameters used for ModernBERT. 

 \subsection{Details for BOW+LogReg experiments}\label{app:bow}
For our bag-of-words logistic regression baseline (Table~\ref{tab:stance_baselines}), we use 4-fold cross validation and report the overall accuracy across all instances across every inference fold. We implement logistic regression using scikit-learn \citep{scikit-learn} with a balanced class weight loss function and an L2 penalty with $C=1.0$. We obtain bag of word representation of the text using unigram tokenizer, removing stop words, setting a minimum document frequency to 5, and the maximum number of features (unigrams) to 1000. 

\begin{table*}[t]
\centering
\resizebox{0.88\linewidth}{!}{
\begin{tabular}{p{0.22\linewidth} p{0.2\linewidth} p{0.22\linewidth} p{0.1\linewidth}}
\toprule
\multicolumn{4}{c}{\textbf{Model and Training Hyperparameters}} \\
\midrule
Dropout & 0.1 & Epochs & 5 \\
Train batch size & 50 & Optimizer & AdamW \\
Learning rate & 3e-5 & Weight decay & 0.05 \\
\midrule
\multicolumn{4}{c}{\textbf{Tokenizer Settings}} \\
\midrule
Padding side & Right & Truncation side & Right \\
Padding strategy & Max length & Truncation strategy & Only first \\
Max sequence length & 512 & & \\
\bottomrule
\end{tabular}
}
\caption{Fixed hyperparameter settings used across all AL simulations and deployment. \label{tab:hyperparams}}
\end{table*}

\section{Acquisition functions}\label{asec:aquisition}

The effectiveness of an active learning pipeline depends heavily on the acquisition function used to select instances for labeling. We evaluated three types of acquisition functions which \citet{zhang2022survey} bucket into categories of informativeness, representativeness, and hybrid.
Methods under informativeness may select challenging but redundant examples, while methods under representativeness may favor easy yet uninformative ones. To address these shortcomings, we also implement contrastive active learning (CAL), a hybrid strategy that tradeoffs both informativeness and representativeness \citep{margatina2021active}. We elaborate on the details of these acquisition functions in the subsections below. 

\subsection{N-Gram Density}

As a baseline acquisition function, we use n-gram density, which selects the next batch of instances for annotation based on unseen n-gram counts \citep{eck2005low}. As a representativeness approach, it prioritizes sentences that introduce new lexical content, encouraging the model to select underrepresented regions of the input space during training. The basic algorithm for a single active learning iteration is as follows:

\begin{enumerate}[itemsep=-4pt, topsep=0pt]
    \item For each unlabeled instance $x_i \in \mathcal{D}_{\text{pool}}$ not already in the sorted list:
    \begin{itemize}[itemsep=-4pt, topsep=-4pt]
        \item Calculate the weight of $x_i$ via Equation~\ref{eqn:n-gram}.
    \end{itemize}
    \item Select the instance with the highest weight and add it to the sorted list.
    \item Repeat until the length of the sorted list reaches the desired batch size $b$.
\end{enumerate}

\vspace{1em}

The set of observed n-grams is cumulative and persists across active learning iterations. The weight of each unlabeled instance $x_i$ is determined by the following equation:

\begin{equation}\label{eqn:n-gram}
\text{weight}_{i,j} = \frac{\sum_{n=1}^{j} \text{count}(\text{unseen } n\text{–grams})}{|\text{sent}|}
\end{equation}

\noindent
where $j$ is the maximum n-gram length considered (e.g., $j = 4$ includes n-grams up to fourgrams);
$\text{count}(\text{unseen } n\text{–grams})$ is the total number of n-grams of length $n$ in the sentence that have not appeared in any previously labeled sentence; and $|\text{sent}|$ is the token length of the sentence.
Repeated occurrences of the same unseen n-gram within a sentence are counted multiple times when computing the weight. 

\subsection{Maximum Entropy}

Maximum entropy is an informativeness acquisition strategy that selects instances for which the model's predictive distribution is most uncertain \citep{schroder2021revisiting}. This approach relies on the output of a supervised model, and selects instances where the predicted class probabilities are closest to uniform, indicating maximal uncertainty. For instance, in binary classification, the largest maximum entropy scores correspond to instances where the predicted probability for each class is close to 0.5.

Formally, for a given unlabeled instance $x_i \in \mathcal{D}_{\text{pool}}$, where $x_i$ is the input and we have $c$ possible classes, we select instances that have the maximum entropy:

\begin{equation}
\arg\max_{x_i} \left[ -\sum_{j=1}^{c} P(y_i = j \mid x_i) \log P(y_i = j \mid x_i) \right]
\end{equation}

\noindent
Here, $P(y_i = j \mid x_i)$ is the model’s predicted probability that instance $x_i$ belongs to class $j$, and the index $i$ ranges over all unlabeled instances in the pool. The acquisition function selects the top $b$ instances with the highest entropy scores.

\subsection{Contrastive Active Learning (CAL)}

Contrastive active learning (CAL) is a hybrid acquisition function that combines informativeness and representativeness by identifying ``contrastive'' instances: unlabeled instances whose model predictions diverge significantly from those of their labeled nearest neighbors (in the model feature space) \citep{margatina2021active}. We re-implement \citeauthor{margatina2021active}'s CAL algorithm, which we report in Algorithm 1.

%\vspace{1em}

\begin{algorithm*}[t!]
\caption{Single iteration of CAL \citet{margatina2021active}}
\textbf{Input:} labeled data $\mathcal{D}_{\text{lab}}$, unlabeled data $\mathcal{D}_{\text{pool}}$, acquisition batch size $b$, model $\mathcal{M}$, number of neighbors $k$, model representation (encoding) function $\Phi(\cdot)$ 
\begin{algorithmic}[1]
\For{$x_p \in \mathcal{D}_{\text{pool}}$}

    \State $\{(x_l^{(i)}, y_l^{(i)})\}, i = 1, \ldots, k \gets \text{KNN}(\Phi(x_p), \Phi(\mathcal{D}_{\text{lab}}), k)$ \hfill \Comment{find neighbors in $\mathcal{D}_{\text{lab}}$}
        
    \State $p(y \mid x_l^{(i)}) \gets \mathcal{M}(x_l^{(i)}), i = 1, \ldots, k$ \hfill \Comment{compute probabilities}

    \State $p(y \mid x_p) \gets \mathcal{M}(x_p)$

    \State $s_{x_p} = \frac{1}{k} \sum_{i=1}^{k} \mathrm{KL}(p(y \mid x_l^{(i)}) \| p(y \mid x_p)), i = 1, \ldots, k$ \hfill \Comment{compute divergence score}
\EndFor
\State \textbf{end}
\State $\mathcal{Q} = \underset{x_p \in \mathcal{D}_{\text{pool}}}{\arg\max} \, s_{x_p}, |\mathcal{Q}| = b$ \hfill \Comment{select batch}
\end{algorithmic}
\textbf{Output:} $\mathcal{Q}$
\end{algorithm*}

In our experiments, we use the \texttt{[CLS]} token embedding of either RoBERTa or ModernBERT as our encoder $\Phi(\cdot)$. For each unlabeled candidate $x_p \in \mathcal{D}_{\text{pool}}$, we identify its $k$ nearest neighbors in the labeled set $\mathcal{D}_{\text{lab}}$, forming a neighborhood: 
\[
\mathcal{N}_{x_p} = \{x_p, x_l^{(1)}, \dots, x_l^{(k)}\}, \quad x_l^{(i)} \in \mathcal{D}_{\text{lab}}
\]

Consistent with the authors' setup, we fix $k = 100$. However, while the authors use Euclidean distance to compute neighborhood similarity, we use cosine distance \citep{jurafsky2025chapter6}.
We then use the model $\mathcal{M}$ to compute the predicted class distributions $p(y \mid x)$ for all $x \in \mathcal{N}_{x_p}$. The final score $s_{x_p}$ for the candidate $x_p$ is computed as the average Kullback--Leibler (KL) divergence between its predicted class distribution and those of its $k$ labeled neighbors.

Finally, the top $b$ instances with the highest scores are selected for labeling in each active learning iteration. Future work could explore alternative hyperparameter choices (e.g., $k=100$, cosine distance) or incorporate manual inspection of nearest neighbors for more nuanced semantic similarity judgments.

\section{Additional AL Simulation Experiments}\label{asec:results}

\subsection{Additional Task for Simulations} \label{asec:al-results}

For AL simulation, we also experiment with the {Trillion Dollar Words (TDW)} dataset which includes sentences from FOMC speeches, meeting minutes, and press conference transcripts \citep{shah2023trillion}. Each sentence is labeled as either \textit{hawkish}, \textit{dovish}, or \textit{neutral}. These labels correspond to whether the sentence advocates for policy tightening (hawkish), easing (dovish), or neither (neutral). 
For example, the following sentence is classified as \emph{dovish}: \emph{``Recent declines in payroll employment and industrial production, while still sizable, were smaller than those registered earlier in 2009''.} In contrast, the following sentence is classified as \emph{hawkish}:  \emph{``Yields on longer-term inflation-indexed Treasury securities, which are relatively illiquid, rose more sharply than did those on nominal securities.''}
Notably, hawkish-dovish classification differs from our desired task because it does not require the stance to be attributed to an individual speaker. Yet, we believe it may be useful for our AL simulations given that the dataset is in the same domain of Federal Reserve communications. 

To match the 25\% positive and 75\% class proportions observed in our prototype annotation rounds, we augmented both the TDW dataset with additional neutral examples. To do so, we use the observation that TDW dataset was originally constructed using a domain-specific keyword filter, and so we randomly sampled sentences from our FOMC transcript corpus that contained keywords the original authors had used as exclusion criteria as our ``neutrals''. For simplicity, we refer to the resulting augmented datasets as TDW+AN, where “AN” denotes the inclusion of augmented neutrals.

\subsection{Full supervision}

In Table~\ref{tab:full_supervision_results}, we report baseline model performance under full supervision, where models are trained on the entire gold-standard training set. This serves as an upper bound on performance for AL methods.

\subsection{Additional Results}

\begin{figure*}[t!]
  \centering
  % modern, 50, vast, 25-75
  \begin{subfigure}[t]{0.48\textwidth}
    \includegraphics[width=\textwidth]{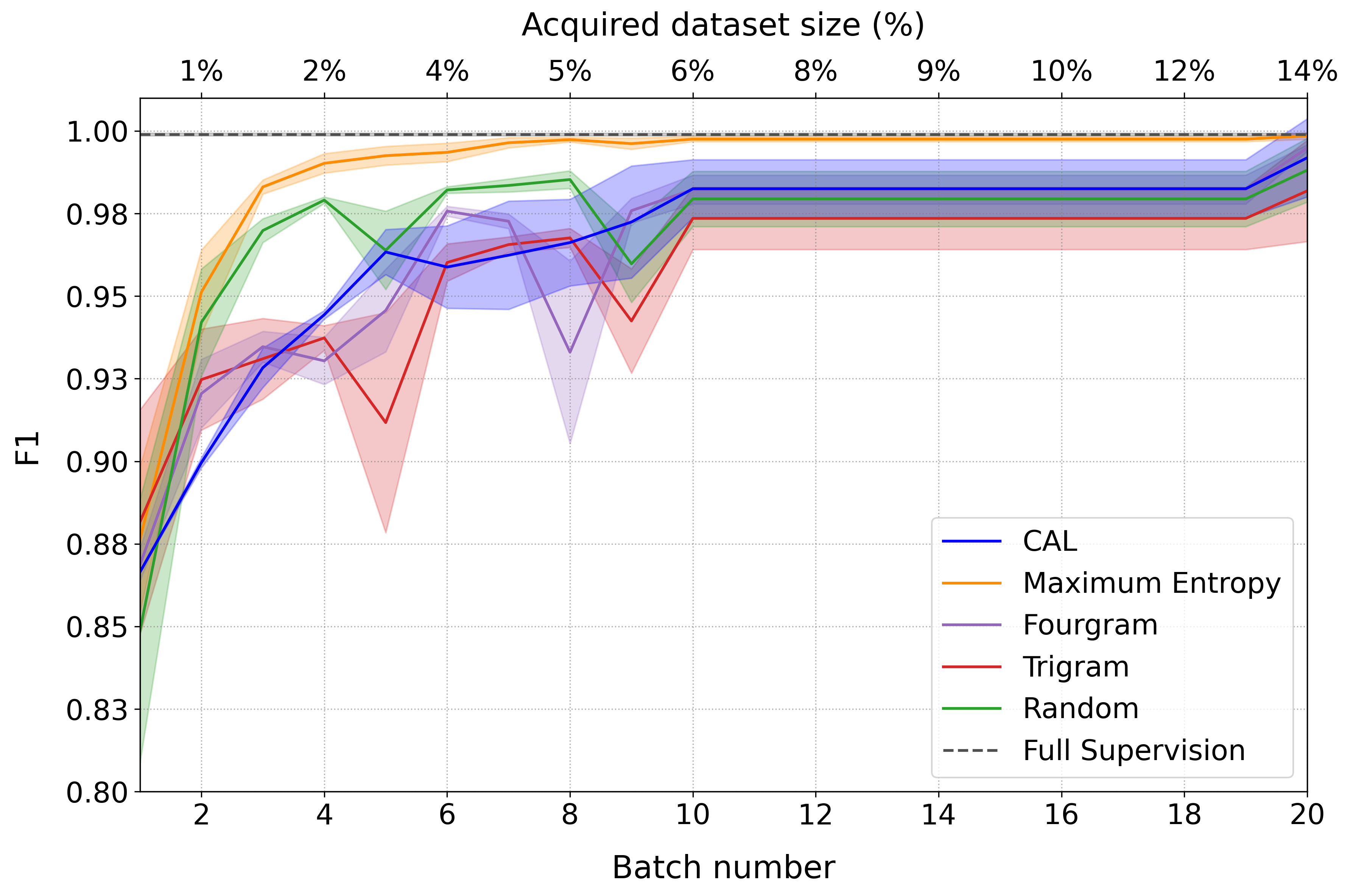}
    \caption{Test set of VAST+AN using ModernBERT, batch size 50, and class imbalance 25\% positive--75\% neutral.}
  \end{subfigure}
  \hfill
  % roberta, 50, vast, 25-75
  \begin{subfigure}[t]{0.48\textwidth}
    \includegraphics[width=\textwidth]{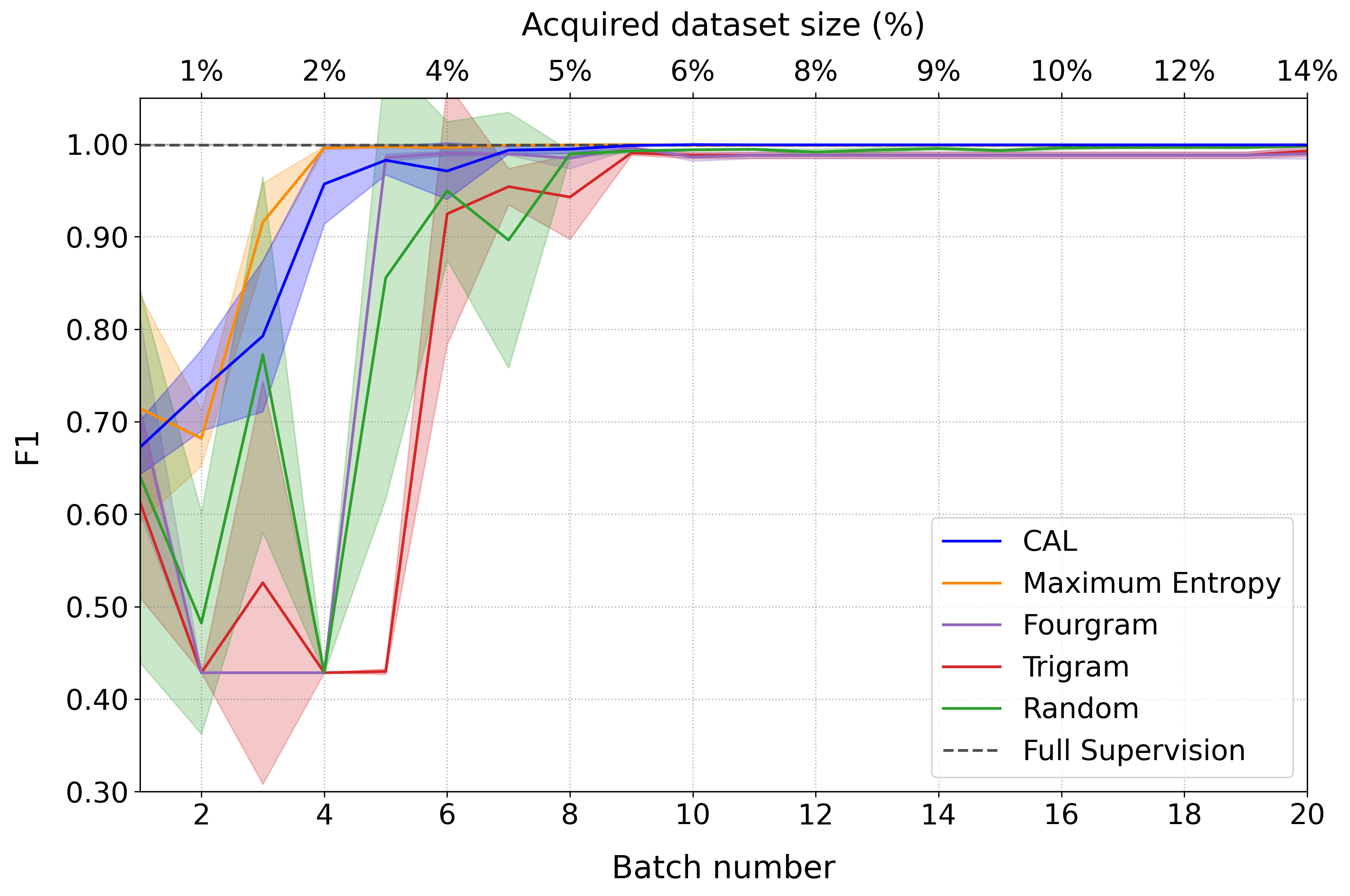}
    \caption{Test set of VAST+AN using RoBERTa, batch size 50, and a class imbalance of 25\% positive--75\% neutral.}
  \end{subfigure}
  % roberta, 150, vast, 25-72
  \begin{subfigure}[t]{0.48\textwidth}
    \includegraphics[width=\textwidth]{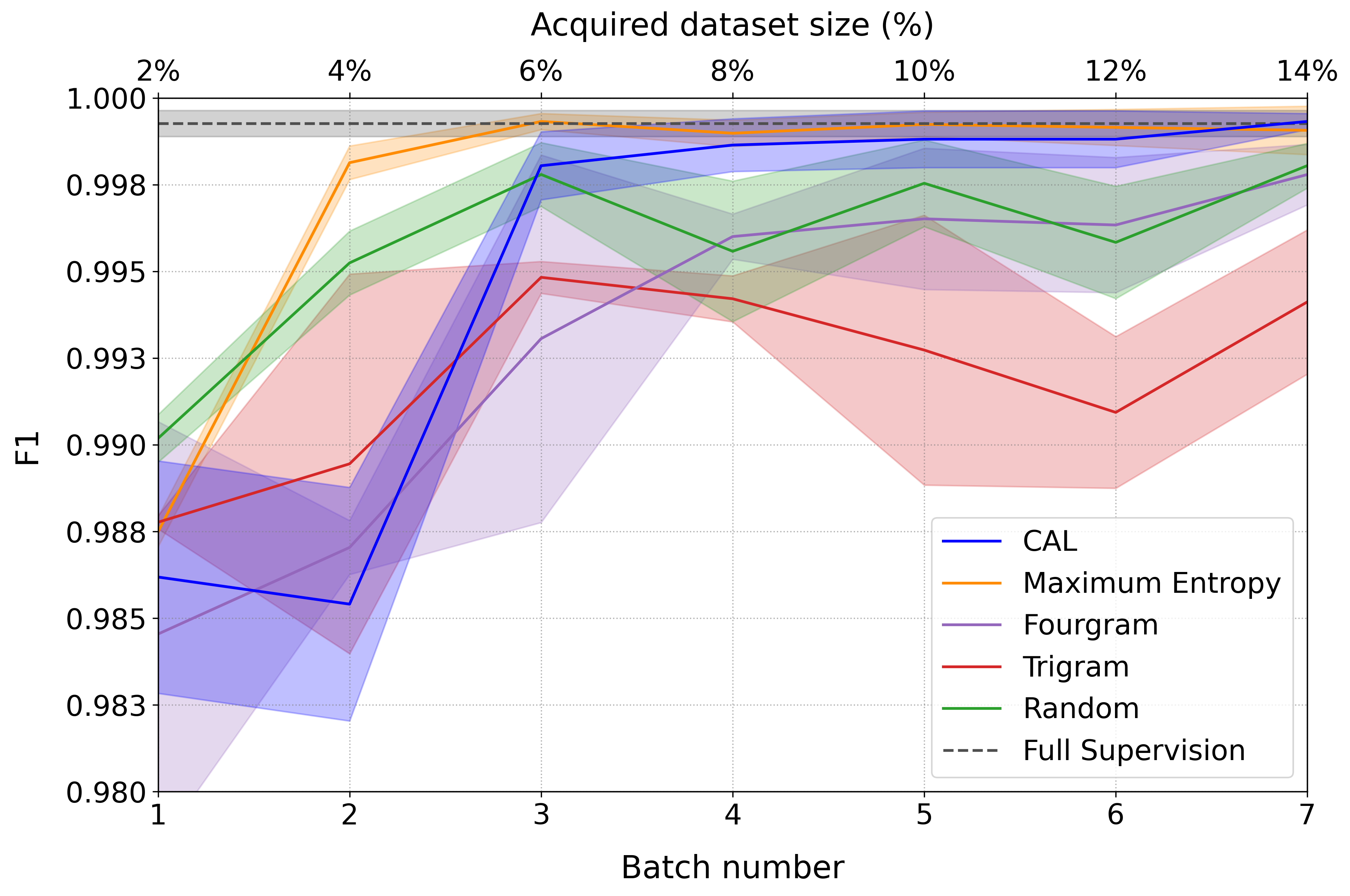}
    \caption{Test set of VAST+AN using RoBERTa, batch size 150, and a class imbalance of 25\% positive--75\% neutral.}
  \end{subfigure}
  \hfill
  % roberta, 50, vast, 10-90
  \begin{subfigure}[t]{0.48\textwidth}
    \includegraphics[width=\textwidth]{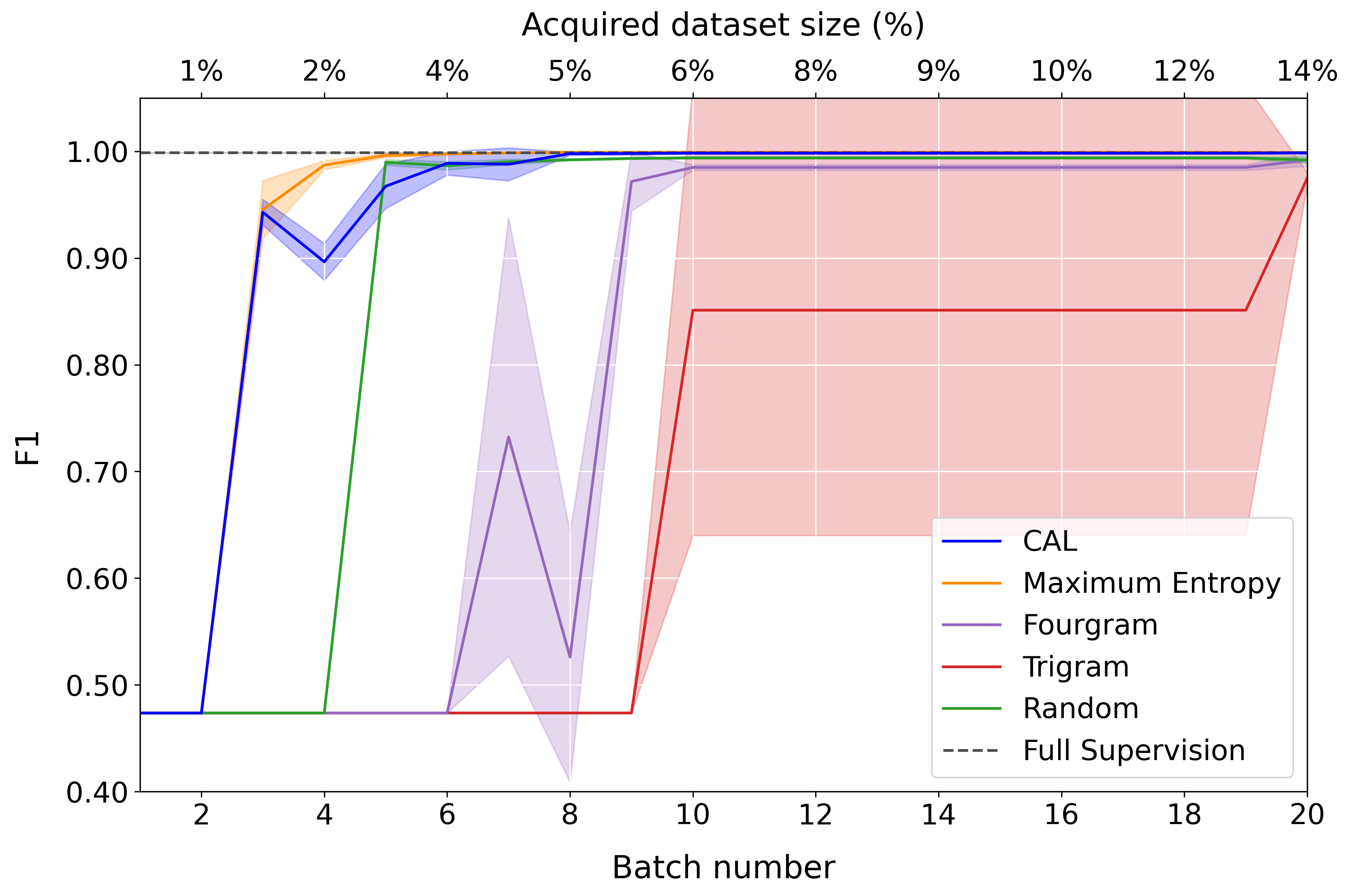}
    \caption{Test set of VAST+AN using RoBERTa, batch size 50, and a class imbalance of 10\% positive--90\% neutral.}
  \end{subfigure}
  % roberta, 50, tdw, 25-75
  \begin{subfigure}[t]{0.48\textwidth}
    \includegraphics[width=\textwidth]{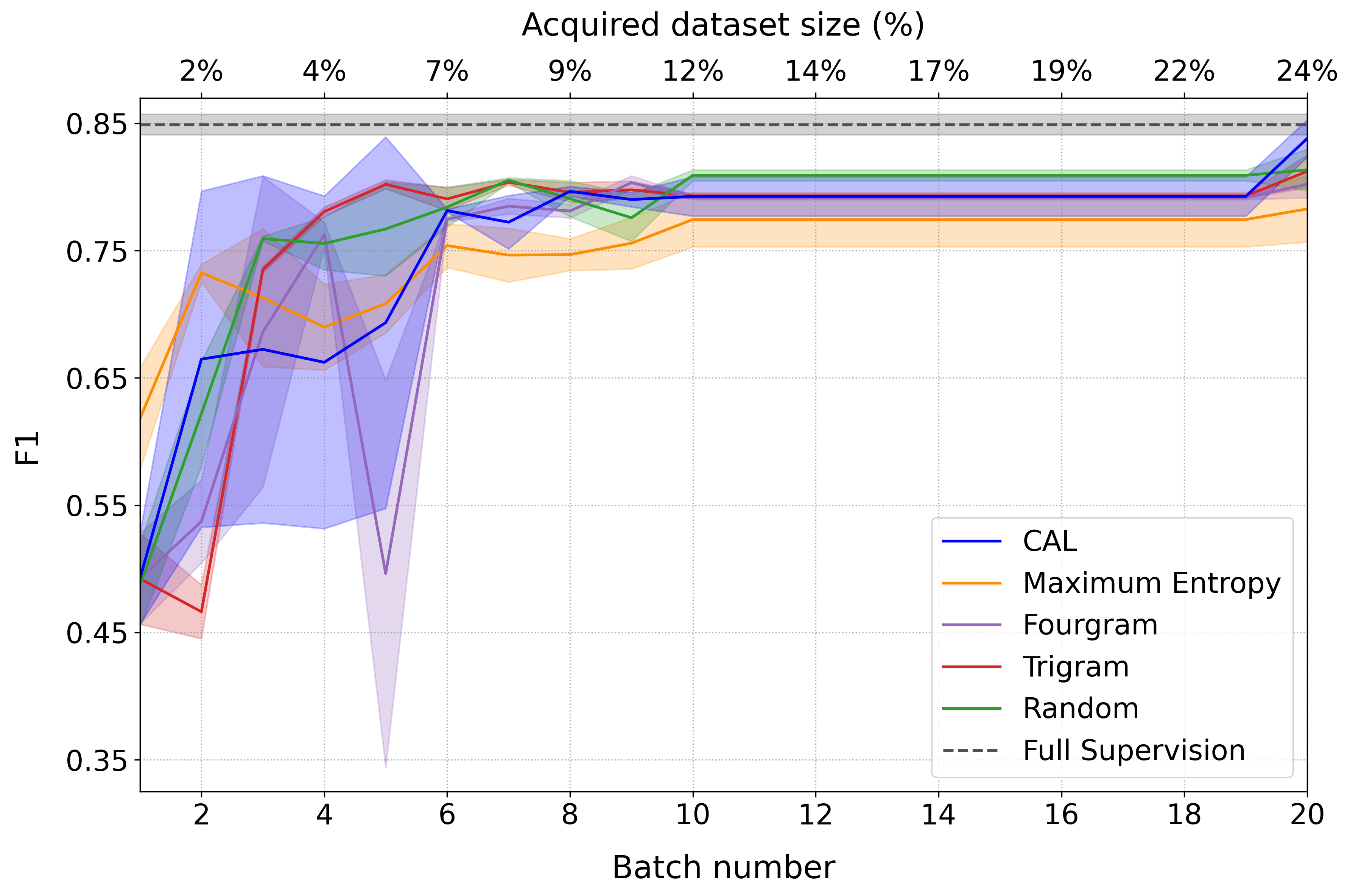}
    \caption{Test set of TDW+AN using RoBERTa, batch size 50, and a class imbalance of 25\% positive--75\% neutral.}
  \end{subfigure}
  \hfill
  \caption{\textbf{Additional active learning simulation results.} We evaluate model performance on the held-out test sets of VAST+AN ($n=3185$) and TDW+AN ($n=4028$) under varying conditions. Each plot reports test macro F1 (y-axis) for models trained on $\mathcal{D}_{\text{lab}}$ at each AL batch (x-axis). Solid lines indicate the mean performance across 5 runs with different random seeds; shaded regions represent the standard deviation. We vary model architecture (ModernBERT vs. RoBERTa), batch size (50 vs. 150), dataset (VAST+AN vs. TDW+AN), and class imbalance (25\%--75\% vs. 10\%--90\%).
  %\kkcomment{Change (b) to be opinion, not stance.}
  \label{fig:three-subfigures}
  }
\end{figure*}

In Figure~\ref{fig:three-subfigures} we show the results of our AL pipeline after varying one decision at a time: the batch size from 50 to 150, the model from ModernBERT to RoBERTa, the simulation dataset from VAST+AN to TDW+AN, and the class distribution from 25\%-75\% to 10\%-90\% (positive-neutral classes respectively). Four out of five of these settings reinforced our choice of using maximum entropy as the acquisition function in our one-shot deployment of AL, as it had an acquisition curve that was steeper or equal to the other methods. 

\begin{table*}[t!]
\centering
\begin{tabular}{lcccc}
\toprule
\textbf{} &
\multicolumn{2}{c}{\textbf{VAST+AN}} &
\multicolumn{2}{c}{\textbf{TDW+AN}} \\
 & \makecell{F1 Train \\ ($n=7459$)} & \makecell{F1 Test \\ ($n=3185$)} & \makecell{F1 Train \\ ($n=4028$)} & \makecell{F1 Test \\ ($n=996$)} \\
\midrule
LogReg + BoW & 0.981 & 0.953 & 0.844 & 0.747 \\
RoBERTa-large & 1.000 & 0.999 & 0.980 & 0.856 \\
ModernBERT-base & 1.000 & 0.999 & 0.998 & 0.826 \\
\bottomrule
\end{tabular}
\caption{Model performance comparison across our gold-standard datasets.}
\label{tab:full_supervision_results}
\end{table*}

\clearpage
\newpage
\clearpage

\section{Prompting Details}\label{appendix:prompts}

In this section, we report the prompt templates as input into the LLM APIs for each of the five tasks from our five-stage hierarchical schema. Each prompt specifies the classification task (using natural language), defines relevant terms, and enforces a labeling constraint. For context-dependent tasks (MP and StanceNLI), the variable \{\$CONTEXT\} refers to the previous 200 tokens (plus any additional tokens required to complete the preceding sentence) followed by the full utterance. This matches the context available to human annotators during the annotation process. In these tasks, the \{\$SENTENCE\} variable also includes the speaker’s name, allowing the model to identify whether the speaker is the same as in the preceding context.

To increase generated answer compliance, we used best practices described by each AI company for each of the LLMs. For the GPT-5 models, we specified a JSON schema (in dictionary form) that listed the valid labels for the \texttt{response\_format} argument. For DeepSeek, we specified an argument for JSON outputs and listed the valid labels in the prompt itself. For Claude, we listed the schemas in their XML format and asked for the response in an XML format.   

textbf{Answer parsing.}
We found the LLMs were 100\% compliant with our constraints of the label output space. So we used \textbf{deterministic answer extraction} and match the surface form of the label with the output of the LLM (since the output is formatted as a JSON object or XML object) \citep{schulhoff2024prompt}.  

\textbf{Prompt templates.}
Below, we provide the prompt templates for DeepSeek since these explicitly listed the label constraints in the prompt. As we mention above, GPT-5 label constraints were passed in via a dictionary and as in an XML format for Claude. 

\newpage
\begin{flushleft}
\textbf{Prompt template for predicting \emph{opinion} (Stage 1)}: 
\\[1em]
\fbox{%
\begin{minipage}{0.9\linewidth}
Is the author expressing an opinion in the sentence below? 
Here, opinion is defined as the expression of a subjective perspective, recommendation, or position.
\\[1em]
Label constraints: ["yes", "no"]
\\[1em]
Output in the following JSON format: \\
"label": your label here, \\
"explanation":  Give a short explanation. \\
"confidence": Include a numerical confidence score between 0 and 1
\\[1em]
Sentence: {\{\$SENTENCE\}}
\\[1em]
JSON output: 
\end{minipage}
}
\end{flushleft}

\begin{flushleft}
\textbf{Prompt for predicting \emph{monetary policy} (Stage 2)}: 
\\[1em]
\fbox{%
\begin{minipage}{0.9\linewidth}
 Does the speaker of the sentence refer to monetary policy, 
            given the context? Here, monetary policy is defined as actions the U.S. Federal
            Reserve may take such as adjusting the federal funds rate, 
            the money supply, or the discount rate.
\\[1em]
Label constraints: ["yes", "no"]
\\[1em]
Output in the following JSON format: \\
"label": your label here, \\
"explanation":  Give a short explanation.\\
"confidence": Include a numerical confidence score between 0 and 1
\\[1em]
Context: {\{\$CONTEXT\}}
\\[1em]
Sentence: {\{\$SENTENCE\}}
\\[1em]
JSON output:
\end{minipage}
}
\end{flushleft}

\newpage
\begin{flushleft}
\textbf{Prompt for monetary policy ``needs context?'' task (Stage 3)}: 
\\[1em]
\fbox{%
\begin{minipage}{0.9\linewidth}
Given the sentence, do you need more context beyond the sentence to determine
whether the sentence is about monetary policy? 
Here, monetary policy is defined as actions the U.S. Federal Reserve 
may take such as adjusting the federal funds rate, the money supply, or the discount rate.
\\[1em]           
Label constraints: ["yes", "no"]
\\[1em]
Output in the following JSON format: \\
"label": your label here, \\
"explanation":  Give a short explanation.\\
"confidence": Include a numerical confidence score between 0 and 1
\\[1em]
Sentence: \{\$SENTENCE\}
\\[1em]
JSON output:
\end{minipage}
}
\end{flushleft}

\begin{flushleft}
\textbf{Prompt for StanceNLI (Stage 4)}: 
\\[1em]
\fbox{
\begin{minipage}{0.9\linewidth}
Given the context, does the sentence entail, contradict, or have a neutral relationship to the hypothesis: 
            "I think we should tighten monetary policy"? 
            Here, tightening monetary policy is defined as actions the U.S. Federal Reserve may take 
            to increase the federal funds rate, such as reducing the money supply or raising the discount rate.\\
\\[1em]            
            Label constraints: ["entailment", "contradiction", "neutral"]
\\[1em]
Output in the following JSON format: \\
"label": your label here, \\
"explanation":  Give a short explanation.\\
"confidence": Include a numerical confidence score between 0 and 1
\\[1em]
Context: \{\$CONTEXT\}
\\[1em]
Sentence: \{\$SENTENCE\}
\\[1em]
JSON Output:
\end{minipage}
}
\end{flushleft}

\newpage
\begin{flushleft}
\textbf{Prompt for StanceNLI ``needs context?'' task (Stage 5)}: 
\\[1em]
\fbox{%
\begin{minipage}{0.9\linewidth}
Given the sentence, do you need more context beyond the sentence to determine
            whether the sentence entails, contradicts, or is unrelated to the hypothesis: 
            ``I think we should tighten monetary policy''? 
            Here, tightening monetary policy is defined as actions the 
            U.S. Federal Reserve may take to increase the federal funds rate, 
            such as reducing the money supply or raising the discount rate.
\\[1em]         
Label constraints: ["yes", "no"]
\\[1em]
Output in the following JSON format: \\
"label": your label here, \\
"explanation":  Give a short explanation.\\
"confidence": Include a numerical confidence score between 0 and 1
\\[1em]
Sentence: \{\$SENTENCE\}
\\[1em]
JSON Output:
\end{minipage}
}
\end{flushleft}

\section{Annotation Codebook}\label{appendix:codebook}

In our data and code repository\footnote{\url{https://github.com/kakeith/op-fed}}, we provide the codebook provided to annotators as part of their training process. The codebook outlines the annotation tasks, label definitions, and examples used to guide consistent and accurate labeling across our five stages. Annotators were instructed to refer back to this document throughout the annotation process. Note, we called Stage 1 ``Stance'' to the annotators but now call Stage 1 ``Opinion'' in the main document due to helpful feedback from a peer reviewer who suggested ``opinion'' more adequately reflected our definition. This relabeling does not change the semantics of the class itself.

% \clearpage
% \newpage

% \includepdf[pages=1-, width=0.95\paperwidth]{attach/codebook.pdf}

\end{document}